\documentclass{article}

\PassOptionsToPackage{numbers, compress}{natbib}
\usepackage[preprint]{neurips_2025}

\usepackage{graphicx}
\usepackage[utf8]{inputenc} % allow utf-8 input
\usepackage[T1]{fontenc}    % use 8-bit T1 fonts
\usepackage{xcolor}
\definecolor{mydarkred}{rgb}{0.6,0,0}
\definecolor{mydarkgreen}{rgb}{0,0.6,0}
\usepackage[colorlinks, linkcolor=mydarkred, citecolor=mydarkgreen]{hyperref}

\usepackage{url}            % simple URL typesetting
\usepackage{booktabs}       % professional-quality tables
\usepackage{amsfonts}       % blackboard math symbols
\usepackage{nicefrac}       % compact symbols for 1/2, etc.
\usepackage{microtype}      % microtypography
\usepackage{xcolor}         % colors
\usepackage{amsmath}
\usepackage{multirow} 
\usepackage{amssymb}
\usepackage{bm}
\usepackage{makecell}
\usepackage{minitoc}
\usepackage{bbm}
\usepackage{subfigure}
\newtheorem{remark}{Remark}
\usepackage{natbib}
\usepackage{wrapfig}
\usepackage{hyperref}

\title{Exploring Imbalanced Annotations for Effective In-Context Learning}

\author{
Hongfu Gao\textsuperscript{1,2}\thanks{Work was done as a research intern at Southern University
of Science and Technology.},\enspace
Feipeng Zhang\textsuperscript{2},\enspace
Hao Zeng\textsuperscript{1},\enspace
Deyu Meng\textsuperscript{3},\enspace
Bingyi Jing\textsuperscript{1},\enspace
Hongxin Wei\textsuperscript{1}\thanks{Corresponding author (\texttt{weihx@sustech.edu.cn})} \\
\textsuperscript{1}Department of Statistics and Data Science, Southern University of Science and Technology \\
\textsuperscript{2}School of Economics and Finance, Xi’an Jiaotong University\\
\textsuperscript{3}School of Mathematics and Statistics, Xi’an Jiaotong University\\
}

\begin{document}
\doparttoc

\maketitle

\begin{abstract}
Large language models (LLMs) have shown impressive performance on downstream tasks through in-context learning (ICL), which heavily relies on the demonstrations selected from annotated datasets. 
However, these datasets often exhibit long-tailed class distributions in real-world scenarios, leading to biased demonstration selection.
In this work, we show that such class imbalances significantly degrade the ICL performance across various tasks, regardless of selection methods.
Moreover, classical rebalancing methods, which focus solely on class weights, yield poor performance due to neglecting condition bias--skewed feature distributions within classes. 
To address this, we propose Reweighting with Conditional Bias (dubbed \textbf{RCB}), a simple and complementary approach to enhance ICL performance under class imbalance. In particular, RCB estimates conditional bias using a balanced subset and re-weights demonstration scores based on both class weight and conditional bias.
In effect, RCB prevents over-selection from dominant classes while preserving the efficacy of current selection methods. 
Extensive experiments on common benchmarks demonstrate the effectiveness of our method, improving the average accuracy of current selection methods by up to 5.42\%.
\end{abstract}

\section{Introduction}
\label{section:introduction}
In-context learning (ICL) has been a remarkable capability of Large Language Models (LLMs), enabling them to perform downstream tasks by simply conditioning on a few task demonstrations. \citep{brown2020language}. 
ICL consistently outperforms zero-shot inference across various tasks without needing parameter updates, positioning it as a strong alternative to supervised fine-tuning (SFT) \citep{mosbach-etal-2023-shot,panwar2024incontext}. 
Previous studies typically focus on designing methods for demonstration selection from high-quality annotated datasets \citep{liu-etal-2022-makes, baldassini2024makes, wang-etal-2024-bayesian}.
The effectiveness of these selection algorithms may hinge on the distribution of annotated datasets, which can often be long-tailed in real-world scenarios \citep{wei2022open, de2024survey} — certain classes are underrepresented (see Figure~\ref{figure:a1}).
Thus, it is of great importance to explore the effect of imbalanced annotations on in-context learning.

In this work, we present the first study on the impact of long-tailed distribution in annotated datasets on in-context learning.
First, we theoretically show that the class prior distribution affects the prediction of ICL.
Empirically, we also find that imbalanced class distributions in annotated datasets significantly degrade the performance of ICL across various tasks, regardless of selection methods (refer to Figure~\ref{figure:1}).
Moreover, traditional rebalancing methods not only fail to ameliorate but, in some cases, exacerbate the issue of class imbalance in ICL.
This motivates us to develop a universal method that consistently improves ICL's performance in the presence of imbalanced annotations.

In this paper, we show that the issue of imbalanced annotations can be mitigated by reweighting the original scores (i.e., cosine similarity score \citep{liu-etal-2022-makes}) of imbalanced examples during selection.
Our method, Reweighting with Conditional Bias (dubbed \textbf{RCB}), is motivated by our analysis of decomposing the distributional differences between imbalanced and test datasets. We find that imbalanced datasets affect the performance of ICL through two lenses: class weights -- the difference between class prior distributions -- and conditional bias, which measures the bias of feature distributions within classes.
The latter is typically neglected in existing rebalancing methods, leading to suboptimal improvements in ICL performance with imbalanced datasets.

Therefore, our key idea behind RCB is to incorporate conditional bias into demonstration reweighting during selection. 
Specifically, we calculate the effective number \citep{cui2019class} as class weights and estimate conditional bias using Bayesian optimization on a balanced subset.
Subsequently, a two-component weight containing both class weights and conditional bias is then used to reweight the demonstration scores for selection.
% These two-component weights are then used to reweight the original scoring functions during selection.
In effect, our approach prevents over-selection from dominant classes while preserving the efficacy of current selection methods.

To verify the effectiveness of our method, we conduct extensive evaluations on seven different downstream datasets, including Amazon \citep{marc_reviews}, AgNews, Yelp, Yahoo \citep{zhang2015character}, Emotion \citep{saravia-etal-2018-carer}, NQ \citep{kwiatkowski-etal-2019-natural}, and CodeSearchNet \citep{husain1909evaluating}. The results demonstrate that our method can largely improve the performance of ICL with the imbalanced datasets. For example, on four classification tasks (Amazon \citep{marc_reviews}, AgNews, Yelp, and Yahoo \citep{zhang2015character}) with a 100 imbalance ratio, our method improves the test accuracy from 46.74$\%$ to 52.16$\%$ – a significant direct improvement of \textbf{5.42}$\%$. Moreover, our approach can generalize to generation tasks to improve ICL's performance with imbalanced annotated datasets. The code and datasets are available in the supplementary material.

Our contributions are summarized as follows:

\begin{enumerate}
    \item We present a phenomenon: imbalanced annotations significantly degrade the performance of ICL regardless of demonstration selection methods. Moreover, classical rebalancing methods fail to address this issue.
    \item We propose a simple and complementary method by involving conditional bias to enhance ICL's performance under class imbalance. Our method is computationally efficient and agnostic to demonstration selection methods.
    \item We empirically validate that our methods can improve the ICL performance in both classification and generation tasks across various imbalance ratios. Our method can be applied to both open-weight LLMs and APIs, as it only requires access to model outputs.
\end{enumerate}

\section{Preliminary}
\subsection{In-context learning}
\label{section:icl}
    In the context of large language models (LLMs), in-context learning (ICL) aims to generate text outputs $\mathrm{\mathbf{y}}=(y_1,...,y_{|\mathrm{\mathbf{y}}|})$ (i.e., token sequences) conditioned on input $\mathrm{\mathbf{x}}=(x_1,...,x_{|\mathrm{\mathbf{x}}|})$ and context $\mathrm{\mathbf{C}}_K$.
    In particular, the context $\mathrm{\mathbf{C}}_K=\{(\mathrm{\mathbf{x}}_i,\mathrm{\mathbf{y}}_i)\}^K_{i=1}$ comprises $K$ task demonstrations (e.g. input-output pairs) selected from a large annotated dataset with $N$ examples $\mathcal{D}_c = \{(\mathrm{\mathbf{x}}_i,\mathrm{\mathbf{y}}_i)\}^N_{i=1}$. 
    Let $f_\theta(\mathrm{\mathbf{C}}_K, \cdot)$ be the ICL model with demonstrations $\mathrm{\mathbf{C}}_K$, using the LLM $f$ parameterized by $\boldsymbol{\theta}$.
    Given a test input $\mathrm{\mathbf{x}}_t$, we generate the output $\mathrm{\mathbf{y}}_t$ via the ICL model as:
\begin{equation}
\label{formula:1}
    \mathrm{\mathbf{y}}_t = f_\theta(\mathrm{\mathbf{C}}_K, \mathrm{\mathbf{x}}_t) =  \mathop{\arg\max}\limits_{\mathrm{\mathbf{y}}} P_\theta  \left(\mathrm{\mathbf{y}}_t \vert \mathrm{\mathbf{C}}_K,\mathrm{\mathbf{x}}_t \right). 
\end{equation}
To improve the performance of ICL, previous studies \citep{liu-etal-2022-makes, peng-etal-2024-revisiting, rubin-etal-2022-learning, Ye2023DPP} designed various scoring functions $s(\cdot, \cdot)$ to select demonstrations $\mathrm{\mathbf{C}}_K$ from an annotated dataset $\mathcal{D}_c$ as:
\begin{align}
\label{formula:2}
\mathrm{\mathbf{C}}_K = \operatorname{Top}_K \left( \left\{ s(\mathrm{\mathbf{c}}_i,\mathrm{\mathbf{x}}_t) \right\}_{i=1}^N \right). 
\end{align}
where $\mathrm{\mathbf{c}}_i$ denotes the $i$-th demonstration and $ \operatorname{Top}_K (\cdot)$ denotes the selection of the $K$ highest-ranked demonstrations from $\mathcal{D}_c$ based on the given scoring function \(s(\cdot, \cdot)\).
For example, Top$K$ \citep{liu-etal-2022-makes} utilizes the cosine similarity distance between $\mathrm{\mathbf{x}}_t$ and $\mathrm{\mathbf{c}}_i$ to select the closest demonstrations.

While current selection methods showcase promising performance on commonly used benchmarks, their effectiveness may hinge on the distribution of annotation datasets $\mathcal{D}_c$. 
For example, ICL might struggle to make accurate predictions for underrepresented groups within these annotated datasets. Real-world datasets (see Figure~\ref{figure:a1}), however, often exhibit an imbalanced distribution, with a few `head' classes containing many examples and numerous `tail' classes having significantly fewer examples. The concern may lead to challenges in effectively employing in-context learning in real-world applications. We proceed with a formulation of the imbalanced setting of ICL.

\subsection{Setting of imbalanced ICL}

In this work, we begin with the class-imbalance setting of in-context learning in classification tasks\footnote{In addition to labels, the imbalance can also occur among various groups, particularly in generation tasks. We extend the imbalanced setting to generation tasks in Section~\ref{section:generation}.
}, where the label space $\mathcal{Y} := \{1, \ldots, k\}$. 
Let $n_j$ denote the number of instances in class $j$, where $j \in \mathcal{Y}$. In the class-imbalanced setting, the annotated dataset $\mathcal{D}_c$ has an unequal distribution of instances across different classes in $\mathcal{Y}$. 
In particular, the class distribution is such that: $n_j \ll n_k$, for some $j,k \in \mathcal{Y}$, where $j \neq k$. 
We quantify the imbalance ratio as $\phi = \frac{\max_{j \in \mathcal{Y}} n_j}{\min_{j \in \mathcal{Y}} n_j}$, and a higher imbalance ratio indicates a more severe class imbalance in the dataset. 

In the real world, class-imbalanced distributions are frequently observed in various datasets. 
For instance, in the Emotion dataset \citep{saravia-etal-2018-carer}, the `Joy' class constitutes 33\% of the data, whereas the `Surprise' class makes up only 4\%.
The CodeSearchNet dataset \citep{husain1909evaluating} includes 29\% JavaScript while Ruby accounts for just 3\%, demonstrating the significant imbalance issue.
Therefore, it is crucial to ensure the ICL performance across all classes under the class-imbalanced setting in $\mathcal{D}_c$.
In what follows, we analyze the effect of class imbalance on ICL from theoretical and empirical perspectives.

\section{Pilot study}

\subsection{The effect of imbalanced data}
\paragraph{Theoretical Analysis} 
In this section, we start by presenting Remark \ref{remark:1} \citep{xie2022an}, which demonstrates that ICL allows LLMs $f$ parameterized by $\theta$ to learn the context-generated distribution from demonstrations $\mathrm{\mathbf{C}}_K$, such that the model's prediction approximates the output $\mathrm{\mathbf{y}}$ generated by the data-generating model $\theta^*$ given the test input $\mathrm{\mathbf{x}}$.

\begin{remark}
\label{remark:1} Assume both demonstrations $\mathrm{\mathbf{C}}_K$ and test sample $(\mathrm{\mathbf{x}}_t, \mathrm{\mathbf{y}}_t)$ are generated by data generated model $\theta^*$. Given such demonstrations $\mathrm{\mathbf{C}}_K$, in-context learning allows large language models to generate output $\mathrm{\mathbf{y}}$ as follows:
\begin{equation}
     f_\theta(\mathrm{\mathbf{C}}_K, \mathrm{\mathbf{x}}_t) \approx \mathop{\arg\max}\limits_{\mathrm{\mathbf{y}}}  P_{\theta^*} (\mathrm{\mathbf{y}}_t|\mathrm{\mathbf{x}}_t). \nonumber
\end{equation}
where large language models $\theta$ to generate correct output $\mathrm{\mathbf{y}}_t$ following context-generated distribution. 
\end{remark}

From a Bayesian perspective, the prediction of ICL is generally made as follows:
\begin{equation}
\label{formula:3}
    f_\theta(\mathrm{\mathbf{C}}_K, \mathrm{\mathbf{x}}_t) \approx   \mathop{\arg\max}\limits_{\mathrm{\mathbf{y}}}  P_{\theta^*} (\mathrm{\mathbf{x}}_t|\mathrm{\mathbf{y}}_t)  P_{\theta^*} (\mathrm{\mathbf{y}}_t). 
\end{equation}
where $P_{\theta^*} (\mathrm{\mathbf{x}}_t)$ and $P_{\theta^*} (\mathrm{\mathbf{y}}_t)$ represent $P_{\theta^*} (X=\mathrm{\mathbf{x}}_t)$ and $P_{\theta^*} (Y=\mathrm{\mathbf{y}}_t)$, respectively. The Eq.~ (\ref{formula:3}) indicates that the class prior distribution $P(Y)$ affects the prediction of ICL.  
For the class-imbalanced setting, the class prior of the test distribution is usually balanced, whereas the annotated dataset exhibits a long-tailed class distribution.
The predicted posterior probability in Eq.~ (\ref{formula:3}) becomes unreliable when we have $P_c(Y=j) \neq P_t(Y=j)$ for any class $j \in \mathcal{Y}$. 
Overall, the above theoretical analysis shows that class-imbalanced annotations cause a mismatch between the class prior distributions of the annotated and test datasets, leading to unreliable posterior probability predictions in ICL. 
Our experimental results further validate the analysis.

\paragraph{Empirical Analysis}

\begin{figure}[t]
\centering  %图片全局居中
\subfigure[]{\includegraphics[width=0.32\textwidth]{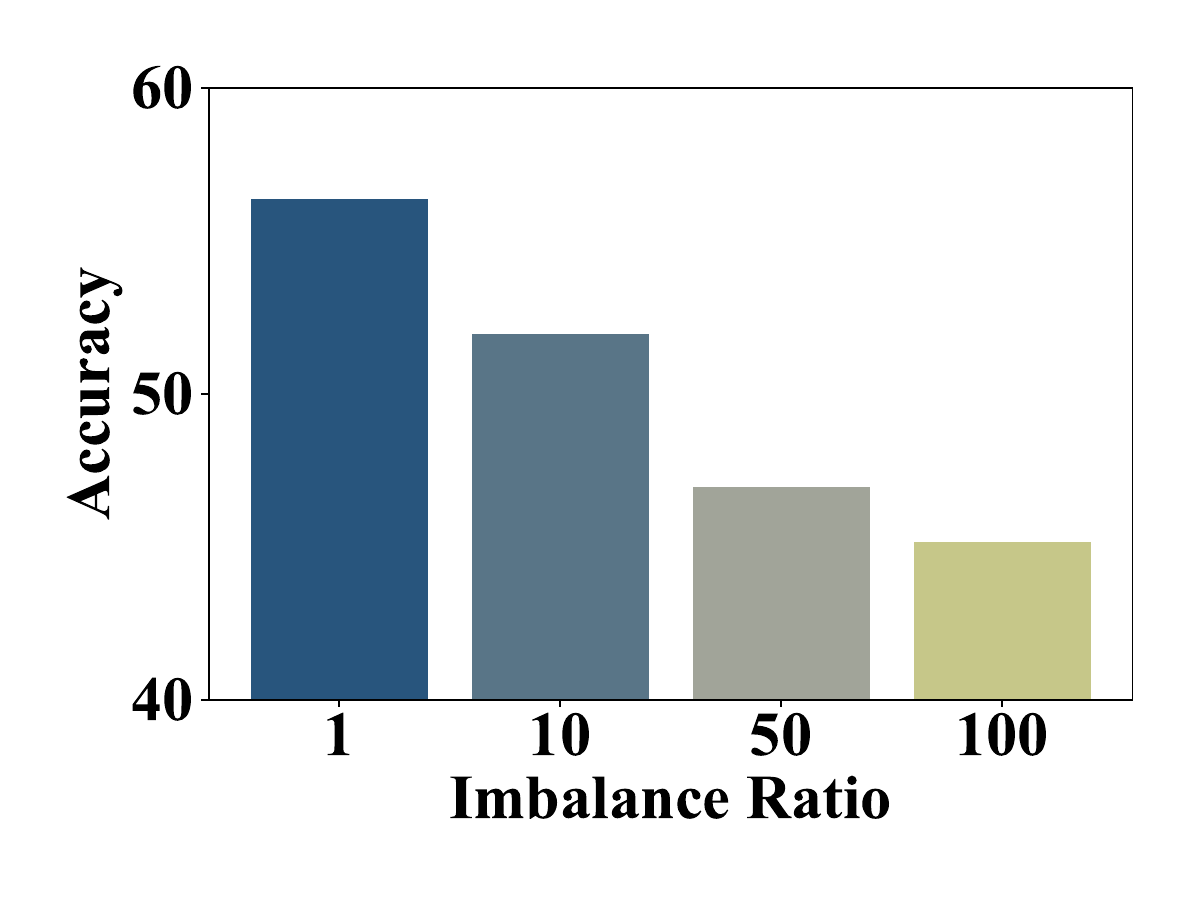}}
\subfigure[]{\includegraphics[width=0.32\textwidth]{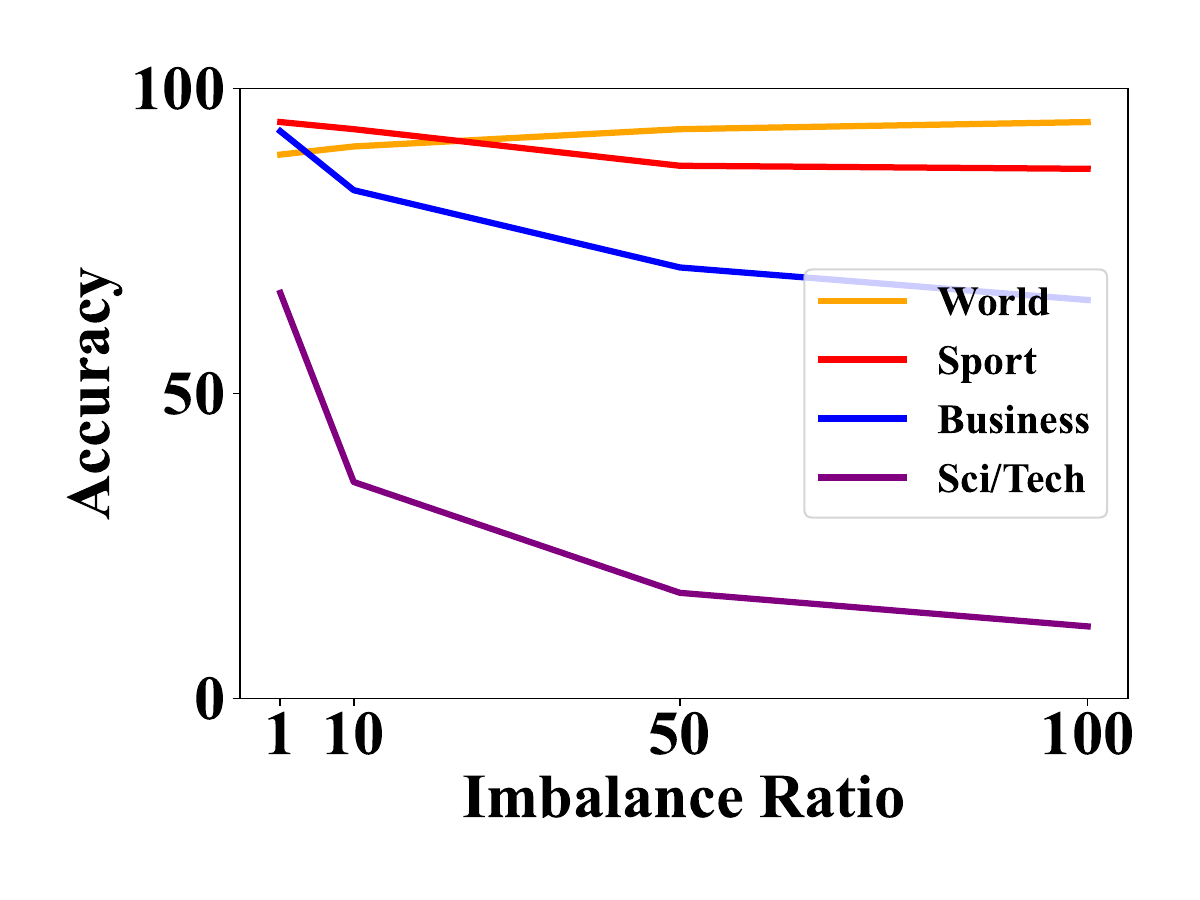}}
\subfigure[]{\includegraphics[width=0.32\textwidth]{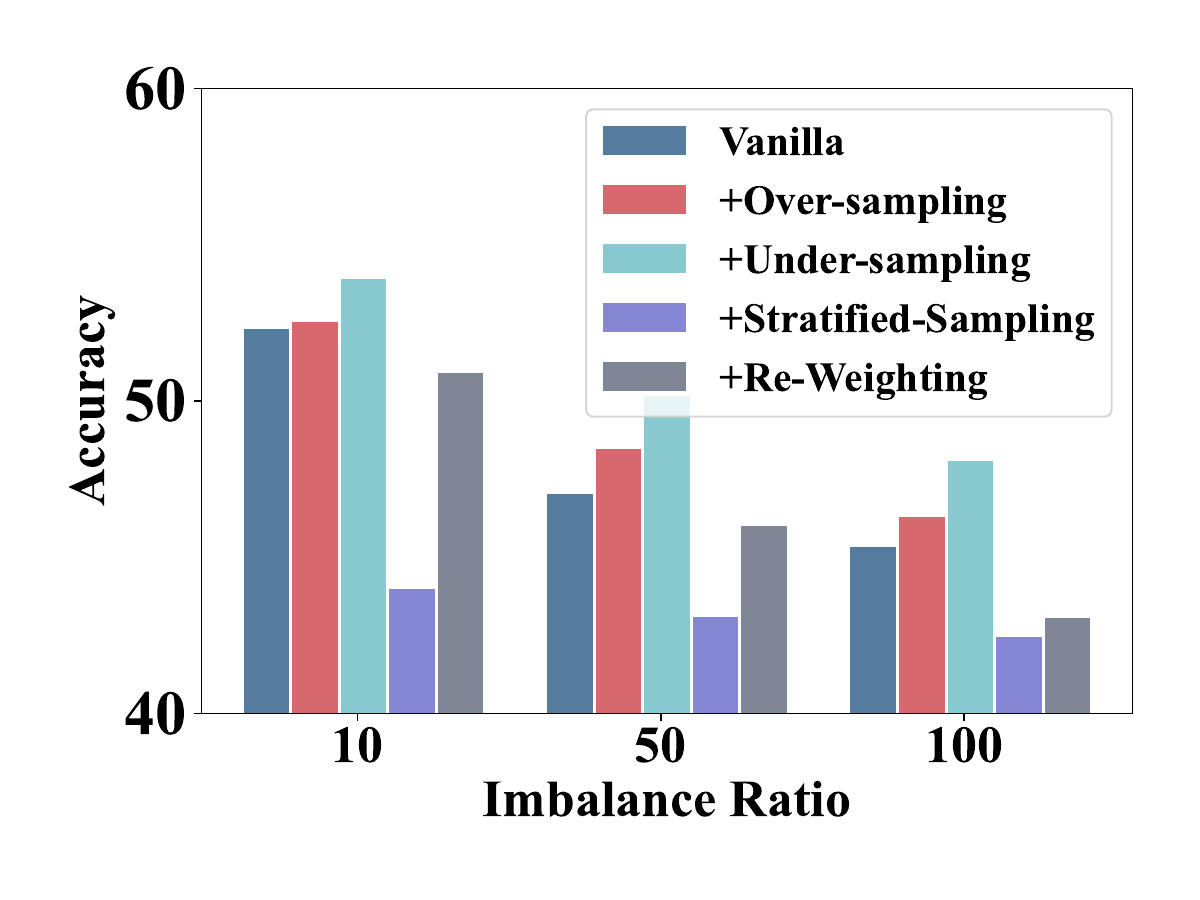}}
\caption{The impact of imbalanced annotations on ICL performance. (a) Overall accuracy across four classification tasks. (b) Average accuracy for each class in AgNews. (c) Overall accuracy across four classification tasks using classical rebalancing methods.}
\label{figure:1}
\vspace{-15pt}
\end{figure}

We also conduct experiments on various downstream tasks, including Amazon \citep{marc_reviews}, AgNews,  Yahoo, Yelp \citep{zhang2015character}, NQ \citep{kwiatkowski-etal-2019-natural}, and CodeSearchNet \citep{husain1909evaluating}). 
To simulate the class-imbalanced setting, we generate imbalanced datasets with a pre-defined probability (e.g.,  $\phi$ = 1, 10, 50, 100).
We evaluate the performance of ICL on a balanced test dataset with various LLMs, including OPT-6.7B, -13B, -30B \citep{zhang2022opt}, LLAMA-3-8B and -70B \citep{llama3modelcard}; and APIs: ChatGPT-3.5-Turbo \citep{achiam2023gpt} and Gemini-2.0-Flash 
 \citep{team2023gemini}.  
We also investigate the performance of ICL with imbalanced annotations using various sizes of demonstrations and selection methods.
% , including Random \citep{min-etal-2022-metaicl}, TopK \citep{liu-etal-2022-makes}, DPP \citep{Ye2023DPP}, VoteK \citep{su2023selective}, ConE \citep{peng-etal-2024-revisiting}, ByCS \citep{wang-etal-2024-bayesian}. 

\textbf{Imbalanced annotations significantly degrade ICL's performance.} Figure \ref{figure:1} (a) shows that selecting demonstrations from an imbalanced dataset significantly deteriorates ICL's performance across various imbalance ratios. 
Specifically, the average accuracy using OPT-6.7B \citep{zhang2022opt} drops approximately \textbf{20}$\%$ for four different classification tasks. 
Figure \ref{figure:1} (b) shows that the decreasing trend in overall accuracy is mainly due to the reduction in accuracy of the tail classes (\textit{Business} and \textit{Sci/Tech}). 
Additionally, the negative effect of imbalanced annotations on ICL is observed in generation tasks, as shown in Section \ref{section:generation}. 
The results of different model architectures and sizes are shown in the Appendix \ref{Appendix_different_model}.  

\textbf{The impact of demonstration number and demonstration selection.}  
Table \ref{table:1} illustrates that selecting a larger set of demonstrations cannot mitigate the issue of imbalanced datasets. 
Meanwhile, the advantages of those powerful selection methods (e.g., TopK \citep{liu-etal-2022-makes} and DPP \citep{ye-etal-2023-complementary}) are neutralized in the presence of an imbalanced dataset. 
For example,  the ICL's performance using the TopK \citep{liu-etal-2022-makes} method decreases from 59.93 to 48.65, a reduction of approximately \textbf{20}$\%$. In conclusion, a larger set of demonstrations or advanced selection methods cannot mitigate the issue of imbalanced annotations.

\subsection{The failure of classical rebalancing methods}
\label{section:rebalanced_method}
One simple and intuitive approach to deal with the class-imbalanced problem is rebalancing \citep{shi2023re}.
In this section, we examine whether four classical rebalancing methods—over-sampling \citep{chawla2002smote}, under-sampling \citep{liu2008exploratory}, stratified sampling \citep{vilarino2005experiments}, and re-weighting \citep{cui2019class}—can mitigate the negative effects of imbalanced annotations on ICL.
Specifically, we select top \textit{K} examples ranked by the score function $s(\cdot, \cdot)$ from an annotated dataset with $N$ examples as demonstrations:
\begin{align}
    \mathrm{\mathbf{C}}_K & =  \operatorname{Top}_K \left(\left\{ s(\mathrm{\mathbf{c}}^{'}_i,\mathrm{\mathbf{x}}_t) \right\}_{i=1}^{N^{'}}\right),\label{formula:5}\\
    \mathrm{\mathbf{C}}_K & = \sum_{i=1}^k\operatorname{top}_{\frac{K}{k}} \left(\left\{ s(\mathrm{\mathbf{c}}_j,\mathrm{\mathbf{x}}_t) \right\}_{j=1}^{n_i}\right),\label{formula:6}\\
    \mathrm{\mathbf{C}}_K & =  \operatorname{Top}_K \left(\left\{w_i\times s(\mathrm{\mathbf{c}}_i,\mathrm{\mathbf{x}}_t) \right\}_{i=1}^N\right).\label{formula:7}
\end{align}
For the over-sampling method in Eq.~(\ref{formula:5}), we select demonstrations $\mathbf{c}^{'}_i$ from an over-sampling dataset with $N^{'}$ examples, where we repeat the examples of tail classes until their number matches that of head classes. 
The under-sampling method in Eq.~(\ref{formula:5}) randomly trims examples belonging to head classes until the number of examples in head classes is equal to that of tail classes.
Suppose $n_j$ represents the size of the class to which the $i$-th example belongs, stratified sampling in Eq.~(\ref{formula:6}) selects $\frac{K}{k}$ demonstrations from each class with $n_j$ examples.
For the re-weighting method in Eq.~(\ref{formula:7}), we select the top $K$ examples based on the scoring functions $s(\mathrm{\mathbf{c}}_i,\mathrm{\mathbf{x}}_t)$ multiplied by the class weights $w_j$ where $w_j = (1-\alpha)/(1-\alpha^{n_j})$ and $\alpha = (N-1)/N$ \citep{cui2019class}.

Figure \ref{figure:1} (c) shows that the rebalancing methods can only achieve limited effects for most downstream tasks, indicating poor generalization to ICL. 
Intuitively, oversampling uses repeated examples that may not provide additional information to LLMs for ICL performance, while undersampling may remove key information of head classes.
For instance, with an imbalanced ratio of 100, over-sampling boosts the performance of ICL from 45.10 to 46.30, yielding limited improvement. 
In contrast, stratified sampling, which selects demonstrations equally from each class, reduces the performance of ICL from 45.10 to 42.46, as it undermines the effectiveness of high-performing selection methods like TopK \citep{liu-etal-2022-makes}.
Notably, we provide an in-depth understanding of the failure through the decomposition of the distributional differences in Subsection \ref{decomping_distributional_difference}.

% Through the empirical analysis, we find that imbalanced annotations significantly hurt the performance of ICL across various datasets, regardless of demonstration selection methods (e.g., TopK \citep{liu-etal-2022-makes} and DPP \citep{ye-etal-2023-complementary}).
% More importantly, employing classical rebalancing methods, like over-sampling \citep{chawla2002smote} and under-sampling \citep{liu-etal-2022-makes}, cannot bridge the gap. 
% This motivates us to design new methods that can universally enhance the performance of ICL in the presence of imbalanced annotations.

\section{Methodology}
\begin{figure}[t]
\centering  %图片全局居中
\subfigure[]{\includegraphics[width=0.32\textwidth]{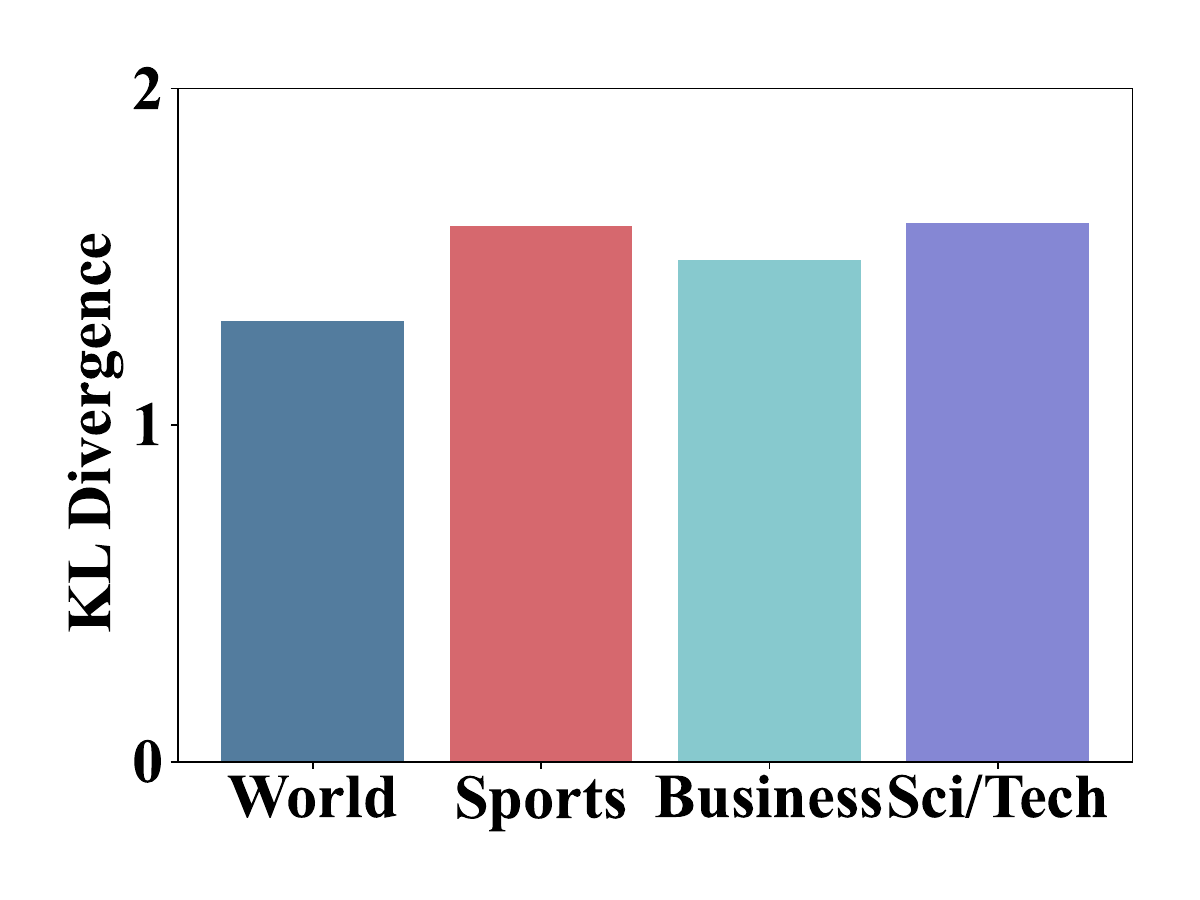}}
\subfigure[]{\includegraphics[width=0.32\textwidth]{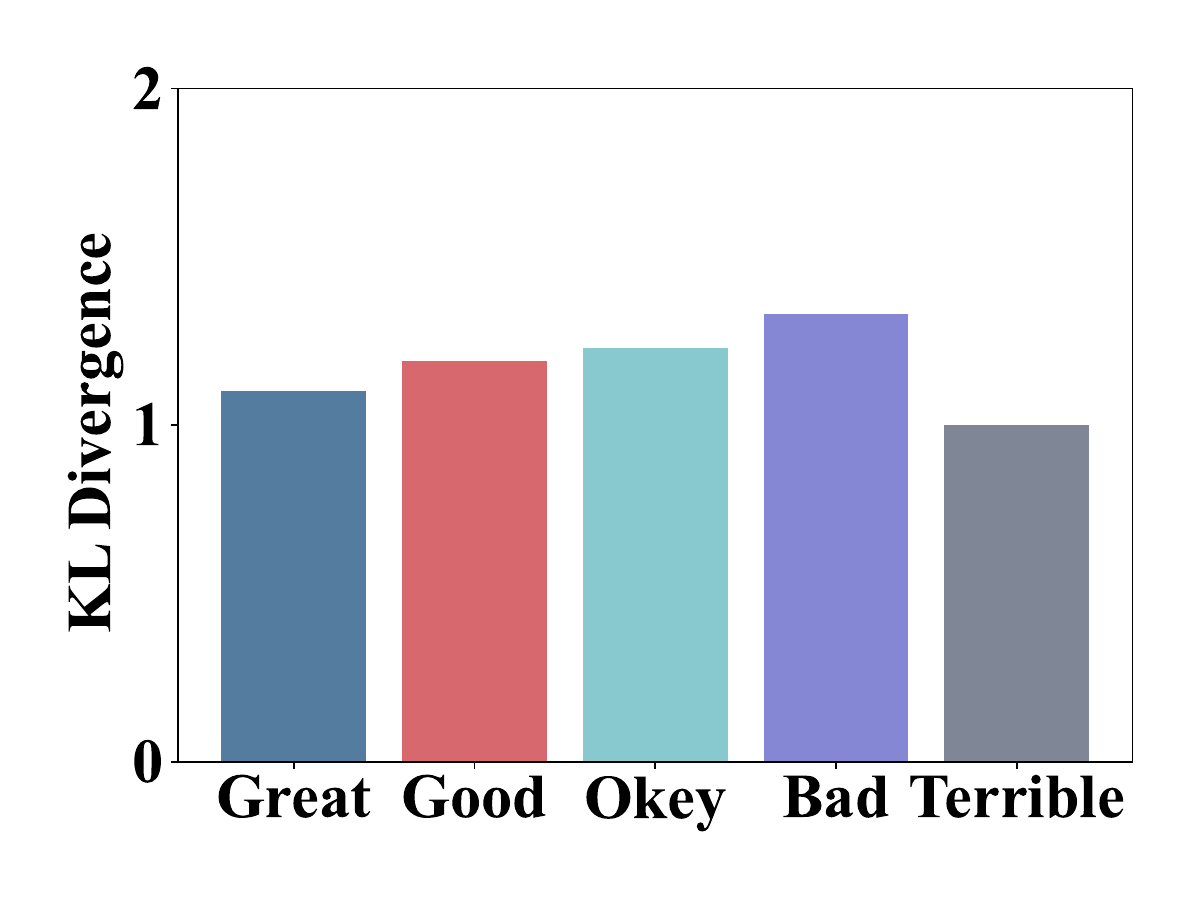}}
\subfigure[]{\includegraphics[width=0.32\textwidth]{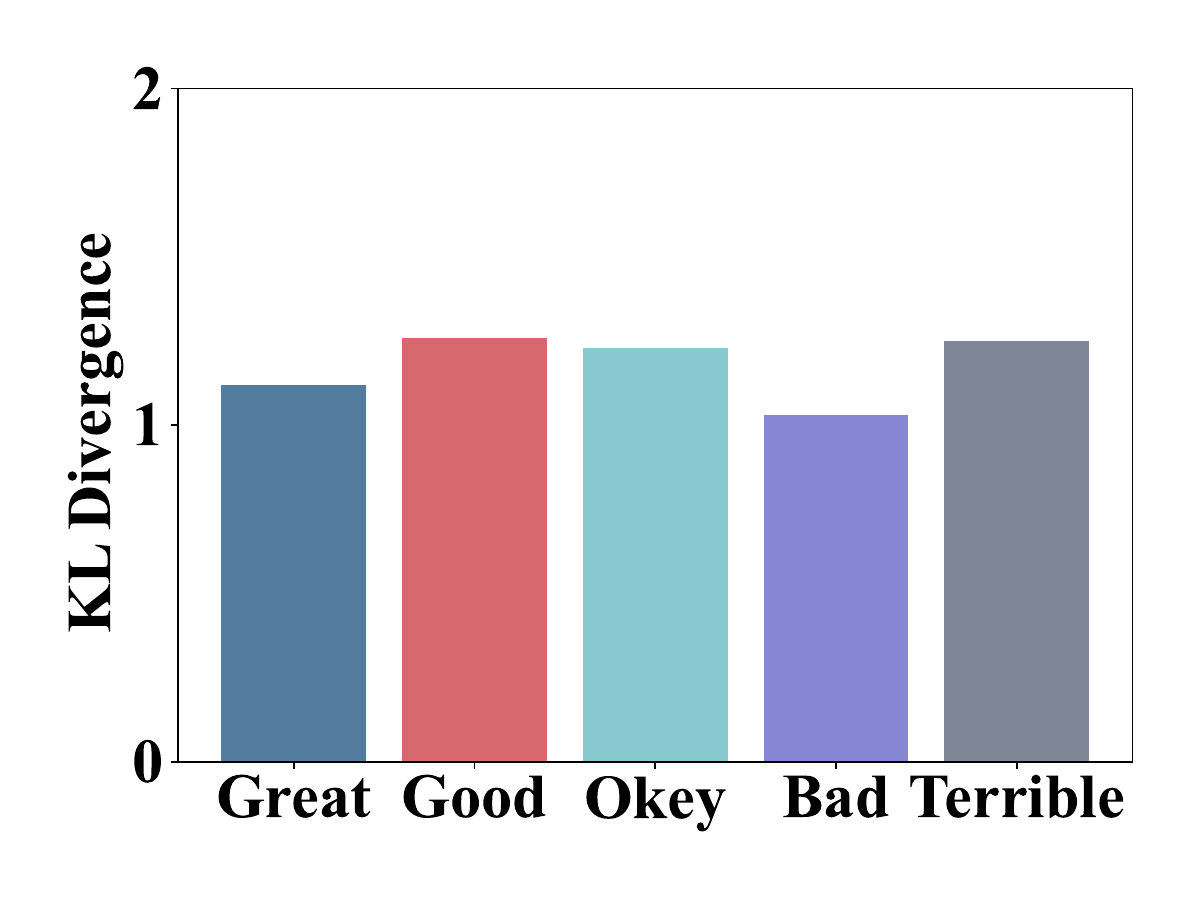}}
\caption{The Kullback-Leibler (KL) divergence between $P_c(X|Y)$ and  $P_t(X|Y)$ for each class in the AgNews (a), Amazon (b) and Yelp (c) datasets. The KL divergence validates the existence of conditional bias in imbalanced datasets.}
\label{figure:2}
\vspace{-15pt}
\end{figure}

\subsection{Decomposition of the distributional differences}
\label{decomping_distributional_difference}
For class-imbalanced annotated datasets $\mathcal{D}_c$, our goal is to select suitable demonstrations for a given test input $\mathrm{\mathbf{x}}_t$ in order to minimize the expected error $\operatorname{Error} $ across all classes:
\begin{align}
    \operatorname{Error} = \mathbb{E}_{P_t(\mathrm{\mathbf{x}},\mathrm{\mathbf{y}})}M\left[ f_\theta\left( \operatorname{Top}_K \left( \left\{ s(\mathrm{\mathbf{c}}_i,\mathrm{\mathbf{x}}_t) \right\}_{i=1}^N \right), \mathrm{\mathbf{x}}_t\right), \mathrm{\mathbf{y}}_t \right],
\end{align}
where $M[\cdot, \cdot]$ measures the mismatched degree between real output and the generated output conditioned on demonstrations $\mathrm{\mathbf{C}}_K$ and test input $\mathrm{\mathbf{x}}_t$. For example, $M[\cdot, \cdot]$ is the error rate for classification tasks and the negative value of the EM score for question-answering tasks.

Next, we apply the importance sampling trick to connect the expected $\operatorname{Error}$ of ICL with the annotated dataset to test dataset as follows:
\begin{align}
\operatorname{Error} &=\mathbb{E}_{P_c(\mathrm{\mathbf{x}},\mathrm{\mathbf{y}})}M\left[ f_\theta\left( \operatorname{Top}_K \left( \left\{ s(\mathrm{\mathbf{c}}_i,\mathrm{\mathbf{x}}_t) \right\}_{i=1}^N \right), \mathrm{\mathbf{x}}_t\right), \mathrm{\mathbf{y}}_t \right] \frac{P_t(\mathrm{\mathbf{x}},\mathrm{\mathbf{y}})}{P_c(\mathrm{\mathbf{x}},\mathrm{\mathbf{y}})}, \nonumber \\
&=\mathbb{E}_{P_c(\mathrm{\mathbf{x}},\mathrm{\mathbf{y}})}M\left[ f_\theta\left( \operatorname{Top}_K \left( \left\{ s(\mathrm{\mathbf{c}}_i,\mathrm{\mathbf{x}}_t) \right\}_{i=1}^N \right), \mathrm{\mathbf{x}}_t\right), \mathrm{\mathbf{y}}_t \right] (\bm{w}+\bm{\beta}). 
    \label{formula:8}
\end{align}
where $P_t(\mathrm{\mathbf{x}},\mathrm{\mathbf{y}})/ P_c(\mathrm{\mathbf{x}},\mathrm{\mathbf{y}})$ measures the distributional difference, $\bm{w} = P_t(Y)/P_c(Y)$ denotes class weights and $\bm{\beta} = \frac{P_t(\mathrm{\mathbf{y}})}{P_c(\mathrm{\mathbf{y}})} \left( \frac{P_t(\mathrm{\mathbf{x}}|\mathrm{\mathbf{y}})}{P_c(\mathrm{\mathbf{x}}|\mathrm{\mathbf{y}})}-1 \right)$ is conditional bias.
The proof is shown in Appendix \ref{section:proof}. 

For an imbalanced dataset, it is easily observed that there is a large discrepancy between the class priors of the annotated and test distributions, resulting in $\bm{w}\neq 1$. 
Due to target shift and concept shift caused by limited sampling in the tailed classes, the conditional distribution between the annotated and test datasets usually shows significant differences, thereby  $\bm{\beta} \neq 0$.
To verify the existence of conditional bias, we employ the Latent Dirichlet Allocation (LDA) model to derive the document-topic distributions for each class and then compute the Kullback-Leibler (KL)  of conditional distribution between the annotated and test datasets. Figure \ref{figure:2} shows that the KL divergence of conditional distribution between the annotated and test datasets is not equal to zero, thereby verifying the existence of conditional bias $\bm{\beta}$ in the imbalanced datasets. Overall, the above analysis illustrates that the performance of ICL is affected by both class weights $\bm{w}$ and conditional bias $\bm{\beta}$; however, the latter is typically neglected in existing rebalancing methods in Subsection \ref{section:rebalanced_method}, leading to suboptimal performance of ICL with imbalanced annotations.

\subsection{Reweighting with Conditional Bias}
\label{section:method}
The previous analysis shows that the performance of ICL significantly deteriorates when demonstrations are selected from an imbalanced dataset without considering the conditional bias $\bm{\beta}$.
However, directly computing the conditional bias $\bm{\beta}$ is non-trivial as real conditional distributions of annotated and test datasets are unknown.
In what follows, we propose \textit{Reweighting with Conditional Bias} (RCB), a complementary strategy to enhance the ICL performance under class imbalance. Our key idea is to incorporate conditional bias into the demonstration reweighting in selection.
With this in mind, we present the details of our approach in the following.

\paragraph{Selecting a balanced subset}
Given an imbalanced dataset $\mathcal{D}_c$ with  $n_j$ examples for the $j$-th class, we select a balanced subset $\mathcal{D}_b$ by uniformly sampling $n_b$ examples from per class of $\mathcal{D}_c$:
\begin{align}
\label{formula:9}
    \mathcal{D}_b = \operatorname{UniformSample}(\mathcal{D}_{c}, n_b),
\end{align}
where $\operatorname{UniformSample}(\mathcal{D}_{c}, n_b)$ denotes uniformly selecting $n_b$ examples per class of $\mathcal{D}_c$, $n_b$ is constrained by the condition $n_b < \min_{j \in \mathcal{Y}} n_j$ to ensure $n_b$ does not exceed the size of the smallest class. 
Through the selection, we construct a balanced subset where all classes contribute an equal number of examples, aligning with the composition of the test dataset. It is worth noting that the selected balanced subset can be also used.

\paragraph{Estimating the conditional bias}
We let the class weights $\bm{w}$ resemble the empirically successful design from previous literature \citep{cui2019class, jamal2020rethinking}. We continue to use \textit{effective numbers} \citep{cui2019class} as class weights $\bm{w}$, defined as $w_j = (1-\alpha)/(1-\alpha^{n_j})$ where $\alpha = (N-1)/N$ and $N$ denotes the total number of examples in imbalanced dataset $\mathcal{D}_c$.
In practice, since the true conditional distribution cannot be directly computed, 
we approximate the conditional bias $\bm{\beta}$ by leveraging the balanced subset $\mathcal{D}_b$:
\begin{align}
\label{formula:11}
\frac{1}{|\mathcal{D}_b|} \sum_{i=1}^{|\mathcal{D}_b|} M\left[ f_\theta\left( \operatorname{Top}_K \left( \{(\bm{w}+\bm{\beta}) \times s(\mathrm{\mathbf{c}}_i,\mathrm{\mathbf{x}}_b) \right\}_{i=1}^{|\mathcal{D}_r|}), \mathrm{\mathbf{x}}_b\right), \mathrm{\mathbf{y}}_b \right],
\end{align}
where we compute the mismatched degree $M$ of the balanced subset $\mathcal{D}_b$ between real output $\bm{y}_b$ and ICL's output conditioned on demonstrations selected from a remaining imbalanced dataset $\mathcal{D}_r = \mathcal{D}_c \setminus \mathcal{D}_b$, while preserving the original imbalance. 
We employ $\operatorname{Top}_K$ to select $K$ highest-ranked demonstrations based on reweighting existing demonstration scores for selection  $s(\cdot, \cdot)$ via $\bm{w}+\bm{\beta}$.
We use a Bayesian optimization framework, which is suitable for non-differentiable and black-box function optimization, to estimate conditional bias $\bm{\beta}$ with the lowest mismatched degree on the balanced subset. 
We show the detailed bayesian optimization process in Appendix \ref{bayeopt}. 
In this way, we can estimate conditional bias $\bm{\beta}$ to measure the bias of feature distributions within classes.

\paragraph{Re-weighting with class weights $\bm{w}$ and conditional bias $\bm{\beta}$} 
Given a test input $\mathrm{\mathbf{x}}_t$, we first select $K^{\prime}$ candidates from the imbalanced dataset $\mathcal{D}_c$ using existing selection methods.
A large  $K^{\prime}$ ensures that all classes (especially tailed classes) are included in the subset of candidates selected. 
For each candidate $\mathbf{c}_i$, we use $\bm{w}+\bm{\beta}$ to reweight its original score $s(\mathrm{\mathbf{c}}_i, \mathrm{\mathbf{x}}_t)$ and employ $\operatorname{argsort}$  to sort the adjusted score in ascending order,  producing sorted indices $\mathcal{I}$:
\begin{gather}
 \mathcal{I} = \operatorname{argsort} \left\{  \right(\bm{w}+\bm{\beta}\left)\times s\right(\mathrm{\mathbf{c}}_i,\mathrm{\mathbf{x}}_t\left)\right\}_{i=1}^{K^{'}}, \nonumber
\end{gather}
We select these candidates with $K$ low-ranking indices in the sorted list $\mathcal{I}$ as the final demonstrations: 
\begin{gather}
    g(\mathrm{\mathbf{c}}_i)  = \mathbbm{1}\left( \operatorname{Loc}(\mathrm{\mathbf{c}}_i, \mathcal{I}) \leq K \right). \nonumber
\end{gather}
where $\mathbbm{1}$ is the indicator function and $\operatorname{Loc}(\mathrm{\mathbf{c}}_i, \mathcal{I})$ return the index of $\mathrm{\mathbf{c}}_i$ in the sorted list $\mathcal{I}$.
After the above steps, we establish the final $K$ demonstration set for in-context learning. 
Noticeably, our method offers several compelling advantages:
\begin{itemize}
    \item \textbf{Algorithm-agnostic}: Our method can be easily incorporated into existing demonstration selection methods, improving the performance of ICL with imbalanced annotations.
    \item  \textbf{Easy to use}: Our method requires access only to the model outputs and integrates effortlessly with any LLMs  (see Figure \ref{figure:3} (d)). Our method is insensitive to the size of the balance subset $|\mathcal{D}_b|$ and class weights $\bm{w}$ (see Figure \ref{figure:3} (a) and (b)). 
     \item  \textbf{High efficiency}: Our method only estimates the conditional bias for each class once for an imbalanced dataset. Thus, our method introduces negligible computational costs compared to in-context learning.  
\end{itemize}

\section{Experiments}
\subsection{Experimental Setup}
\label{section:experimental_setting}
\textbf{Datasets and Evaluation} 
For our evaluations,  we verify the effectiveness of our method on four benchmark datasets, including Sentiment Classification (Amazon \citep{marc_reviews}, Yelp \citep{zhang2015character}) and  Topic Classification (AgNews, Yahoo \citep{zhang2015character}). 
Due to limited space, these tasks' input/output, statistics and split are reported in the Appendix \ref{appendix:experimental_setting}. 
To simulate the issue of imbalanced data, we generate imbalanced datasets with pre-defined imbalance ratios $\phi$ (e.g., 1, 10, 50, 100) and we evaluate the performance of ICL on another corresponding class-balanced test dataset \citep{shi2023re,jamal2020rethinking}. 
We report average accuracy (\%) with standard deviation (over 3 runs).

\textbf{Models and Baselines} We utilize various large language models, including open-weight models: OPT-6.7B, OPT-13B, OPT-30B, LLAMA-3-8B and LLAMA-3-70B; and APIs: ChatGPT-3.5-Turbo and Gemini-2.0-Flash.
We use Bert-base-uncased sentence encoder \citep{Bert} as the similarity tokenizer. 
We also conduct experiments with existing selection methods as baselines, including \textbf{Random} \citep{min-etal-2022-metaicl}, \textbf{TopK} \citep{liu-etal-2022-makes}, \textbf{DPP} \citep{Ye2023DPP}, \textbf{VoteK} \citep{su2023selective}, \textbf{ConE} \citep{peng-etal-2024-revisiting}, \textbf{ByCS} \citep{wang-etal-2024-bayesian}.  
For hyperparameters, we set the size of the balanced subset as $|\mathcal{D}_b| =100$ and $K^{\prime}=1600$ by default.
The details of our experiments are presented in Appendix \ref{appendix:experimental_setting}.
\begin{table}[!t]
\caption{
Average test accuracy ($\%$) with standard deviation of various selection methods on four classification datasets with various imbalanced ratios (over 3 runs). Bold numbers are superior results. Our strategy consistently enhance the performance of existing methods.}
    \centering
    \resizebox{1\columnwidth}{!}{%
    \begin{tabular}{ccccccccc}
    \toprule
         $K$&  \multicolumn{4}{c}{8}&  \multicolumn{4}{c}{16}\\
\toprule
         \textbf{Imbalance Ratio}&  \textbf{1}&  \textbf{10}&  \textbf{50}&  \textbf{100} &  \textbf{1}&  \textbf{10}&  \textbf{50}& \textbf{100} \\
\midrule
Random&  47.70$\pm$0.77&  43.72$\pm$1.26&  39.59$\pm$0.92&  38.01$\pm$0.58&  50.68$\pm$0.49&  46.65$\pm$0.72&  41.30$\pm$0.99& 
39.90$\pm$0.54\\
 \textbf{+Ours}& 47.62$\pm$1.06& \textbf{47.81$\pm$1.40}& \textbf{46.51$\pm$0.90}& \textbf{47.50$\pm$1.13}&50.11$\pm$0.70& \textbf{50.50$\pm$0.60}& \textbf{50.41$\pm$1.35}&\textbf{51.01$\pm$2.19}
\\
 TopK& 58.32$\pm$0.03& 54.01$\pm$0.68& 48.52$\pm$0.99& 46.78$\pm$1.50& 59.93$\pm$0.05& 55.57$\pm$0.58& 50.78$\pm$0.88&
48.65$\pm$0.88
\\
\textbf{+Ours}&58.12$\pm$0.65& \textbf{57.46$\pm$0.50}& \textbf{54.14$\pm$1.31}& \textbf{51.92$\pm$1.22}& 59.53$\pm$0.37& \textbf{58.94$\pm$0.93}& \textbf{55.58$\pm$0.75}&\textbf{52.83$\pm$1.32}
\\
 DPP& 58.50$\pm$0.00& 53.90$\pm$0.79& 49.37$\pm$0.78& 47.25$\pm$1.11& 60.00$\pm$0.07& 55.54$\pm$0.85& 51.30$\pm$0.74&
49.58$\pm$0.58
\\
\textbf{+Ours}& 58.20$\pm$0.34& \textbf{56.76$\pm$1.09}& \textbf{53.25$\pm$1.32}& \textbf{51.93$\pm$1.13}& 59.90$\pm$0.11& \textbf{58.40$\pm$0.94}& \textbf{55.06$\pm$1.10}&\textbf{53.34$\pm$1.26}
\\
 VoteK& 58.07$\pm$0.14& 53.51$\pm$0.78& 48.75$\pm$0.92& 47.21$\pm$1.26& 60.00$\pm$0.15& 55.53$\pm$0.49& 50.46$\pm$0.79&
48.42$\pm$0.90\\
  \textbf{+Ours}& 57.92$\pm$0.30&\textbf{56.20$\pm$0.68}& \textbf{53.95$\pm$1.34}& \textbf{51.84$\pm$1.01}&59.89$\pm$0.22& \textbf{58.23$\pm$0.81}& \textbf{55.45$\pm$0.63}&\textbf{52.93$\pm$0.89}
\\
 ConE& 57.36$\pm$0.51& 52.41$\pm$1.09& 47.86$\pm$1.33& 45.58$\pm$1.33& 58.60$\pm$0.51& 53.97$\pm$0.71& 49.25$\pm$0.70&46.87$\pm$0.80\\
 \textbf{+Ours}& 57.59$\pm$0.62& \textbf{54.75$\pm$0.91}& \textbf{52.45$\pm$0.70}& \textbf{50.12$\pm$1.29}& 58.85$\pm$0.41& \textbf{56.06$\pm$0.81}& \textbf{54.19$\pm$0.74}&\textbf{51.17$\pm$0.75}
\\
 ByDC& 58.38$\pm$0.40& 52.62$\pm$0.87& 47.95$\pm$1.09& 45.78$\pm$1.17& 60.04$\pm$0.31& 53.91$\pm$0.54& 49.66$\pm$0.77&47.01$\pm$0.71
\\
\textbf{+Ours}&  58.16$\pm$0.37&  \textbf{57.24$\pm$0.78}& \textbf{53.25$\pm$1.27}&  \textbf{51.45$\pm$0.93}&  60.03$\pm$0.28&  \textbf{57.53$\pm$0.89}& \textbf{54.53$\pm$0.70}& \textbf{51.44$\pm$0.79}
\\\hline 
\multirow{2}{*}{Average}&56.39$\pm$0.31&51.69$\pm$0.91&47.01$\pm$1.00&45.10$\pm$1.16&58.21$\pm$0.26&53.53$\pm$0.65&48.79$\pm$0.82& 46.74$\pm$0.73
\\
&56.27$\pm$0.56&\textbf{55.04$\pm$0.89}&\textbf{52.26$\pm$1.14}&\textbf{50.79$\pm$1.12}&58.05$\pm$0.35&\textbf{56.61$\pm$0.83}&\textbf{54.08$\pm$0.88}& \textbf{52.16$\pm$1.20}\\\hline
    \end{tabular}
    }
    \label{table:1}
    \vspace{-10pt}
\end{table}

\subsection{Main Results} 
\textbf{Can our method improve existing ICL methods against imbalanced annotations?} Table \ref{table:1} presents the average accuracy of ICL using different selection methods on four benchmark datasets with OPT-6.7B, under various imbalance ratios.
A salient observation is that our method drastically improves the performance of ICL with imbalanced annotated datasets. 
For example, for the 100 imbalance ratio, our method improves the average accuracy of six existing selection methods from 46.74$\%$ to 52.16$\%$ – a \textbf{5.42} of direct improvement.  
More importantly, we show that our method can boost performance for a wide range of selection methods, such as TopK and DPP.  
For example, we observe that, for the 100 imbalance ratio, the test accuracy of ICL using TopK is improved to \textbf{4.18} when integrating our methods. 
Table \ref{table:1} also presents that our method can consistently improve the performance of ICL against imbalanced datasets with varying demonstration numbers $K$. 
The detailed results of each dataset are reported in Appendix \ref{appendix_more_results}.

\begin{figure}[t]
\centering % 确保整个figure*环境中的内容居中  
\subfigure[]{\includegraphics[width=0.24\textwidth]{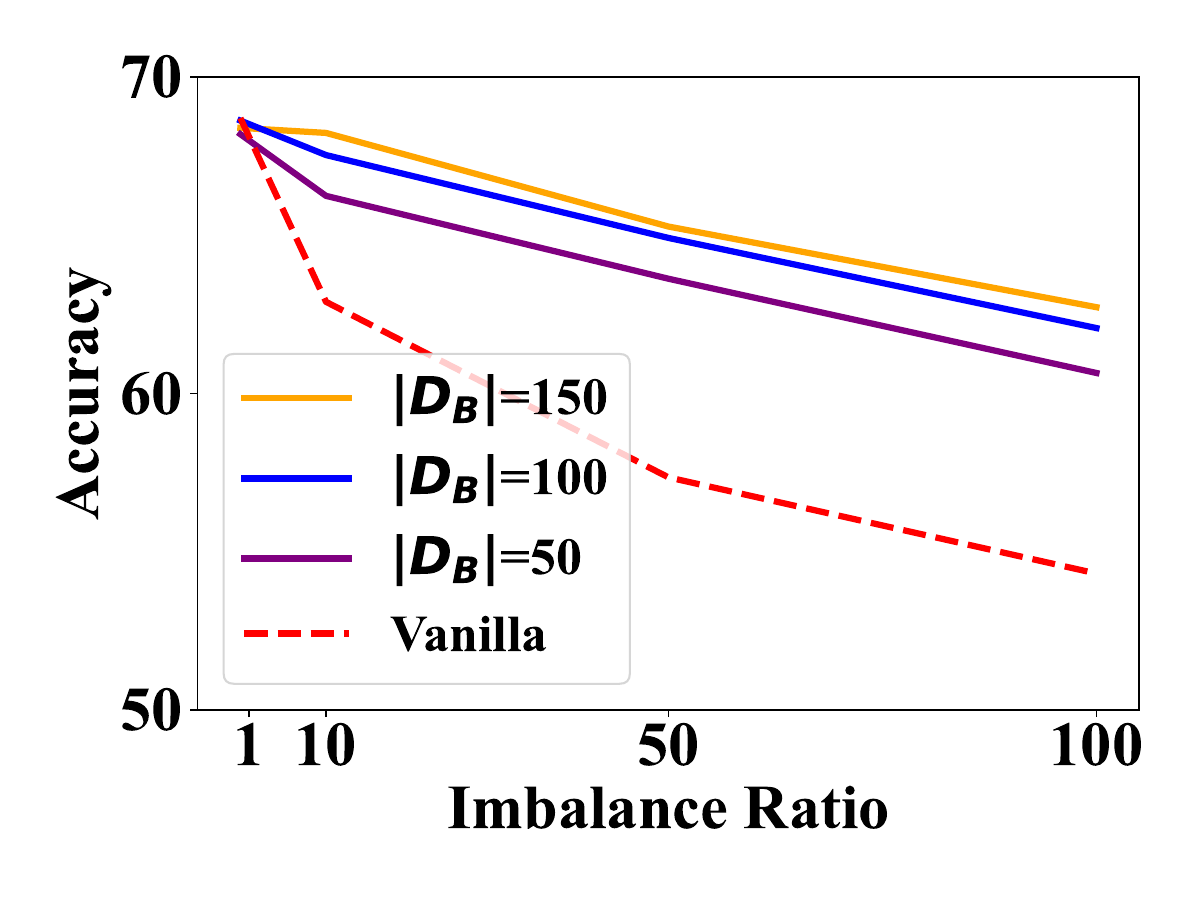}} % 
\subfigure[]{\includegraphics[width=0.24\textwidth]{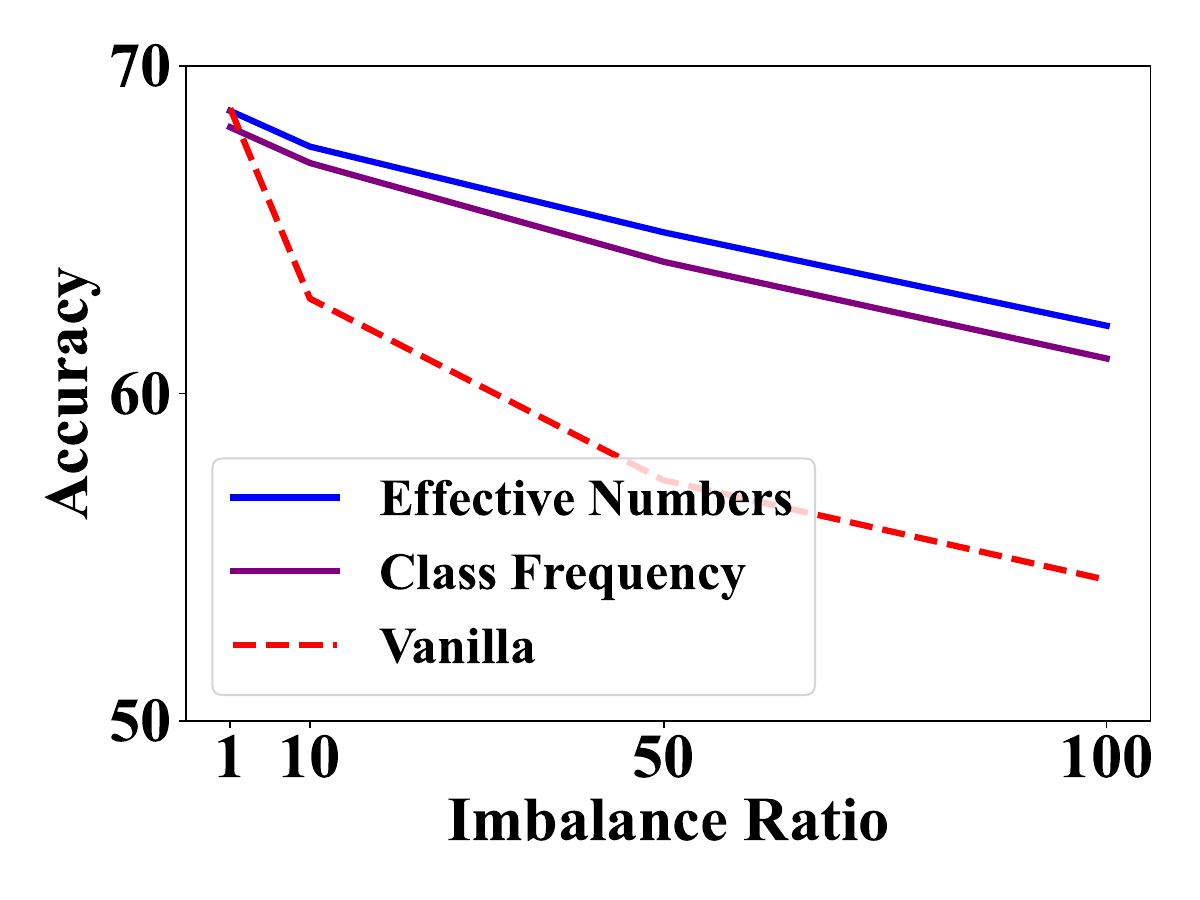}} %
\subfigure[]{\includegraphics[width=0.24\textwidth]{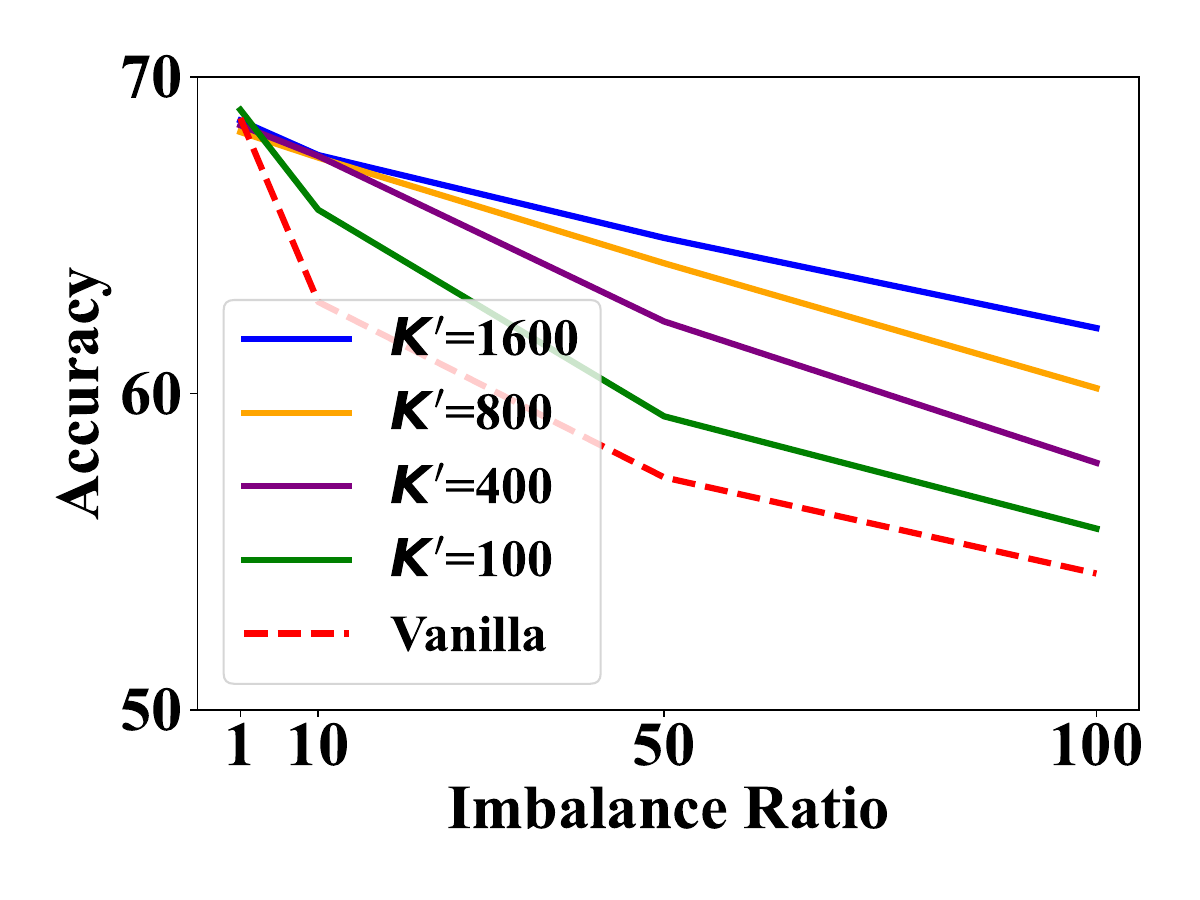}} % 
\subfigure[]{\includegraphics[width=0.24\textwidth]{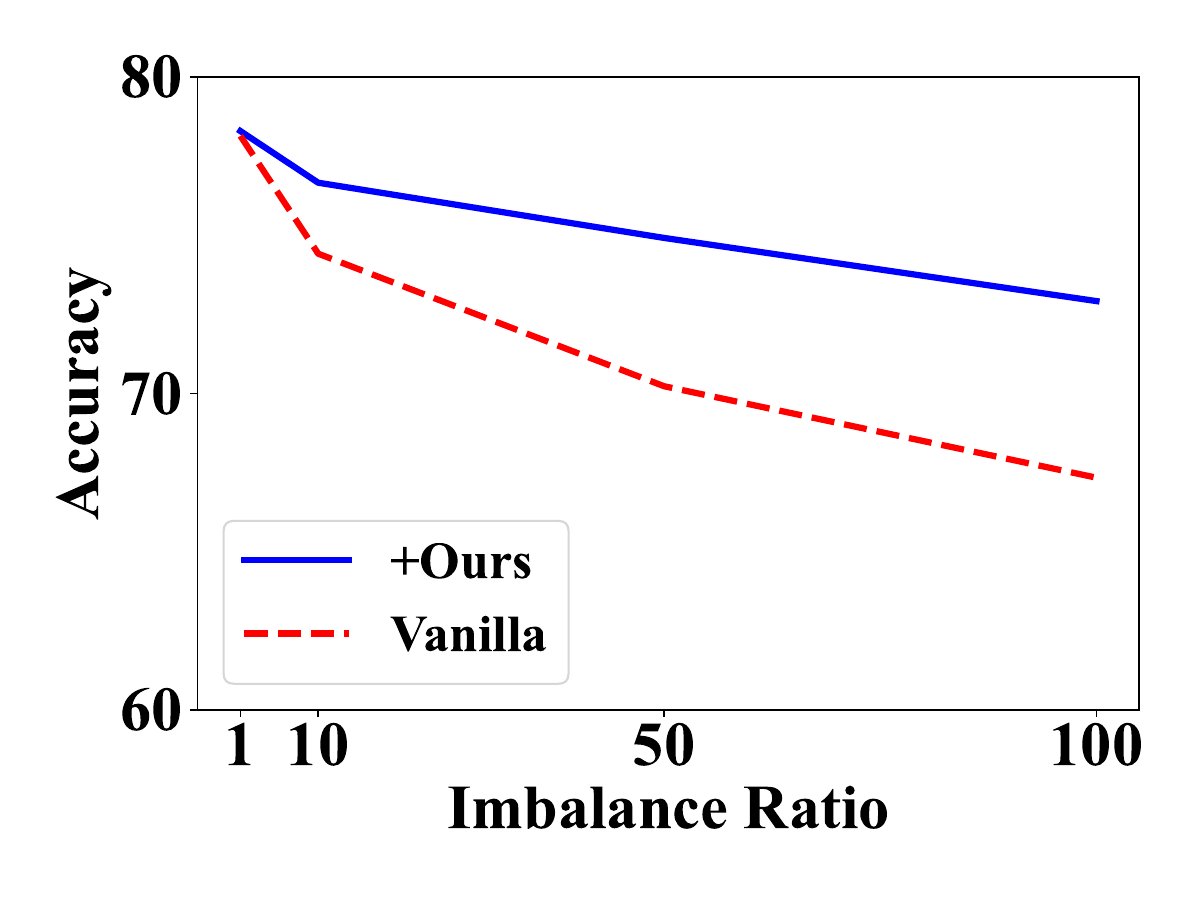}} % 
\caption{Results for ablation studies. (a)  Ablation on the different balanced subset sizes. (b) Ablation on the different class weights (c) Ablation on the different selected candidates. (d) Ablation on the different model architectures and sizes.}
\label{figure:3}
\vspace{-10pt}
\end{figure}

\textbf{How does the balanced subset affect the robustness of our method?} 
In Figure \ref{figure:3} (a), we examine how the size of the balanced subset $\mathcal{D}_b$ affects the effectiveness of our method (cf. Eq. \ref{formula:9}). 
The \textit{Vanilla} indicates all candidate demonstrations are selected without our method. 
It’s noteworthy that our method shows robustness to the choice of the size of the balanced subset $|\mathcal{D}_b|$. 
Even when we set $|\mathcal{D}_b| = 50$, it still yields significant improvements in ICL performance on AgNews and Yahoo datasets across six selection methods against imbalanced datasets.
Notably, the balanced subset can be reused in the process of demonstration selection. Thus, our method is applicable in scenarios with extremely limited data from tail classes.

\textbf{Is our method robust with different class weights $\bm{w}$?} 
In the paper, we employ \textit{effective numbers} as class weights $\bm{w}$, which have been confirmed to successfully deal with imbalance problems in previous studies \citep{cui2019class,jamal2020rethinking}. 
We employ \textit{class frequency} $w_j=\frac{n_j}{N}$ \citep{shi2023re,Kang2020Decoupling}  as an alternative class weights to verify whether the type of class weights affects the performance of our methods.  Figure \ref{figure:3} (b)  shows that our method with \textit{class frequency} can improve the ICL's performance across various imbalance ratios and achieve accuracy similar to that of \textit{effective numbers}.

\textbf{Does our method work with the different number selected candidates $K^{\prime}$?} 
We set a large value of $K^{\prime}$ to ensure that all classes (especially tailed classes) are included in the subset of candidates selected by existing methods. 
Given a small $K^{\prime}$, the candidate subset may contain too few examples from tailed classes (even missing), leading to suboptimal performance. Therefore, the $K^{\prime}$ should be larger in cases of larger imbalanced ratios. 
We validate the claim by conducting experiments with varying $K^{\prime}$ on datasets with various imbalanced ratios. The result in Figure \ref{figure:3} (c) demonstrates that one should set a sufficiently large $K^{\prime}$ to achieve the best performance. 

\textbf{Is our method effective with different model architectures and sizes?} 
To show our proposed method is model-agnostic, we conduct experiments on a diverse collection of model architectures and sizes, including open-weight models: OPT-6.7B, OPT-13B, OPT-30B, LLAMA-3-8B and LLAMA-3-70B; and APIs: ChatGPT-3.5-Turbo and Gemini-2.0-Flash.  
The results (vanilla/+ours) in the Figure \ref{figure:3} (d) present the same phenomenon as the main experiments in the manuscript: the ICL performance of LLMs gets worse at larger imbalanced ratios, and our method can significantly improve the performance, especially at large imbalanced ratios (e.g., 100). 
For instance, for LLAMA-3-8B \citep{llama3modelcard}, using our method boosts the average accuracy from 65.72 to 73.20, a 7.48 of direct improvement with a 100 imbalance ratio.
The detailed results of each model are presented in the Table \ref{table:a5}.

\textbf{Can our method work in real-world imbalanced datasets?}
We verify the effectiveness of our method on a real-world imbalanced dataset. 
In Emotion \citep{saravia-etal-2018-carer},  a few sentiments make a large contribution and data tend to show a long-tailed distribution (refer to Figure \ref{figure:a1})..  For example, ``joy`` and ``sadness`` are head classes, while most other classes, such as ``love`` and ``surprise``, are tail classes.  Table \ref{table:a8} shows that our method consistently improves existing selection methods and outperforms existing rebalancing methods. 
For example, with OPT-6.7B \citep{zhang2022opt}, using our method boosts the average Macro-F1 of six selection methods from 32.12 to 36.69, a direct improvement of 4.57. 
These results verify that our method is effective in improving ICL's performance in real-world imbalanced scenarios.

% In conclusion, we verify the effectiveness of our method for mitigating the issue of imbalanced annotations across several benchmark datasets with multiple existing selection methods. We also apply our methods to improve the performance of ICL against imbalanced annotations in real-world scenarios and find it a consistently effective solution.

\section{Discussion}
\subsection{Data imbalance in text generation tasks}
\label{section:generation}
\begin{figure}[t]
\centering  %图片全局居中
\subfigure[]{\includegraphics[width=0.32\textwidth]{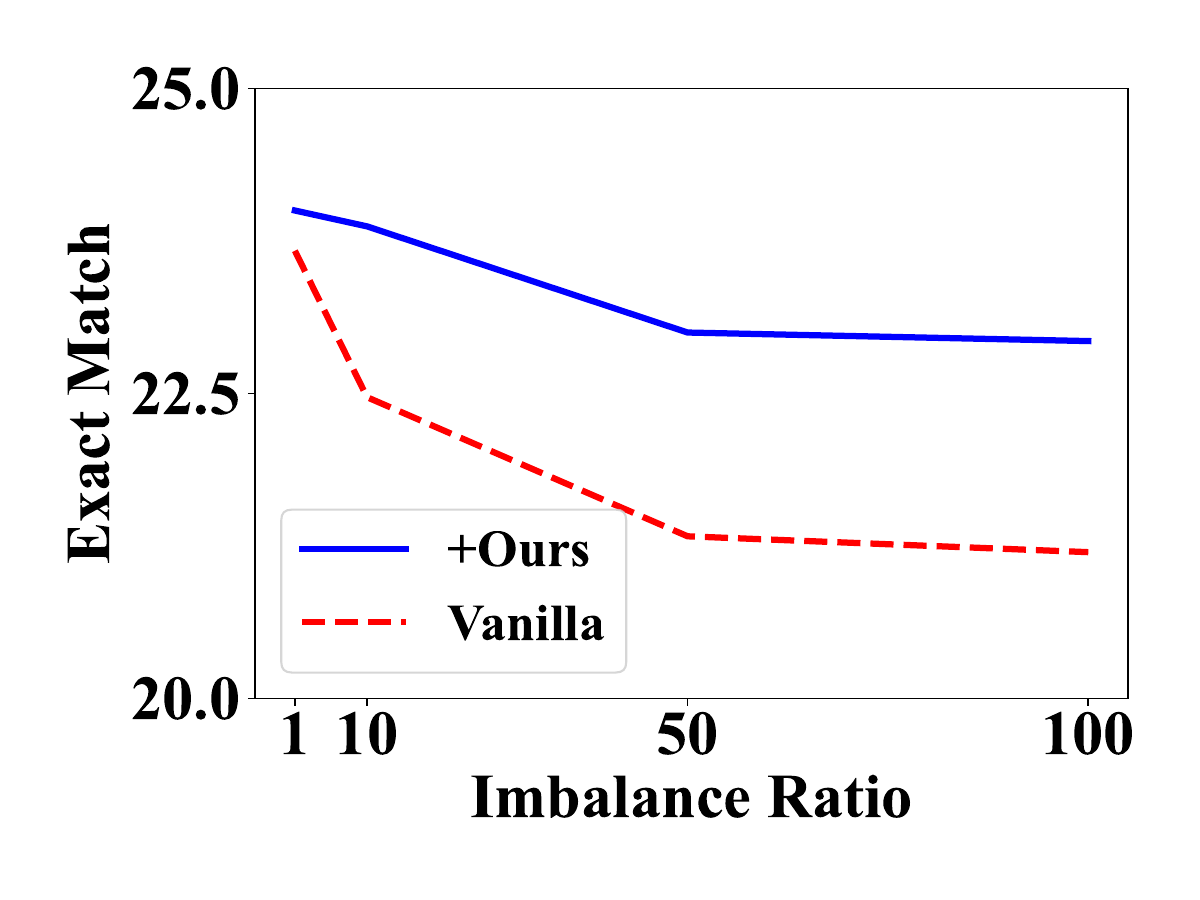}}
\subfigure[]{\includegraphics[width=0.32\textwidth]{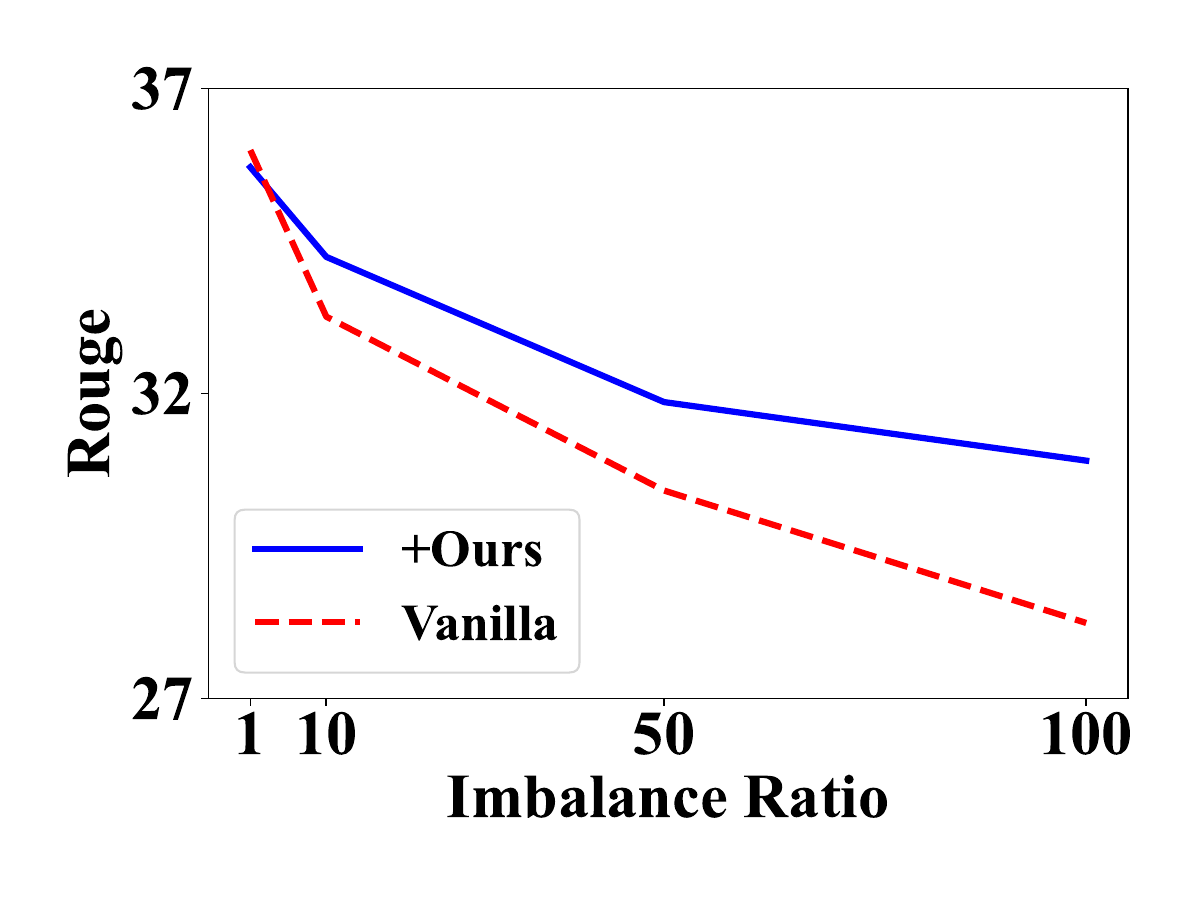}}
\subfigure[]{\includegraphics[width=0.32\textwidth]{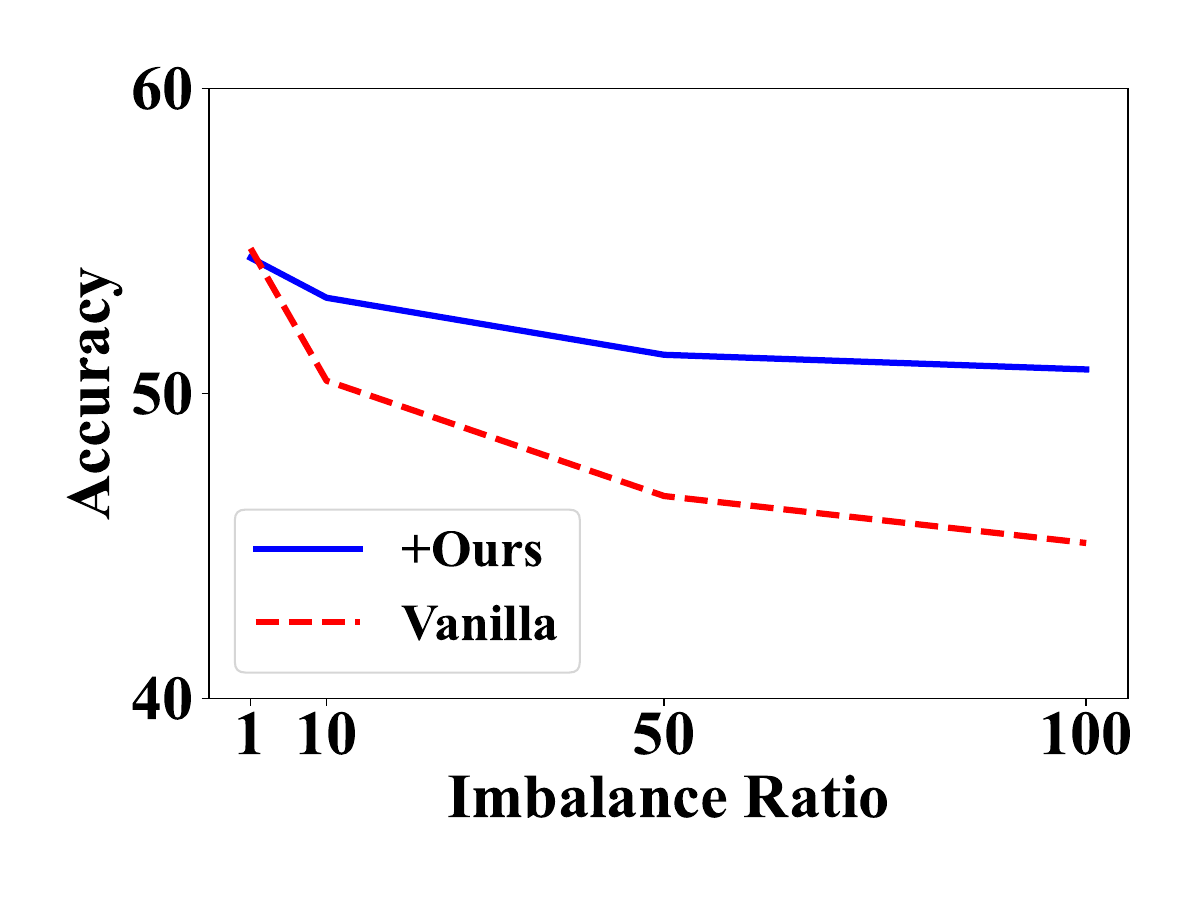}}
\caption{The impact of imbalanced annotations on the performance of In-Context Learning (ICL) in generation tasks, including Natural Questions (NQ) (a) and CodeSearchNet (b). Figure (c) shows the result of imbalanced datasets with an increasing number of examples on four classification tasks.}
\label{figure:4}
\vspace{-15pt}
\end{figure}

Text generation, which is a common task in in-context learning, may also follow a long-tailed distribution.
To address this, we verify the effect of imbalanced annotated datasets and the effectiveness of the proposed method in text generation tasks. 
Specifically, we consider two generation tasks: Open-Domain Question-Answering (NQ \citep{kwiatkowski-etal-2019-natural}) and Code Summarization (CodeSearchNet \citep{husain1909evaluating}).
The NQ \citep{kwiatkowski-etal-2019-natural} dataset can be divided into five categories, including person (10.41$\%$), time (20.32$\%$), geography (9.02$\%$), culture (45.18$\%$), and professional knowledge (15.06$\%$).
Figure \ref{figure:4} (a) and (b) report the average performance of NQ and  CodeSearchNet using six existing selection methods across various imbalance ratios.

Figure \ref{figure:4} (a) and (b) demonstrate that imbalanced annotation significantly hurts ICL's performance on generation tasks. 
We also observe that our method consistently improves the test performance on NQ and  CodeSearchNet, indicating the generalization of our method to generation tasks. 
For example, our method improves the average EM score of NQ with an imbalance ratio of 100 from 21.20 to 22.93, a relative improvement of 8.16$\%$ compared to vanilla methods.

\subsection{Does a larger dataset address the imbalance issue?}
While our analysis has shown that imbalanced annotations negatively impact ICL by reducing the number of examples in tail classes, one might wonder \textit{if this negative effect is due to the reduced size of the dataset rather than a shift in class prior distribution}. 
To address this, we verify the negative impact of imbalanced annotations and the effectiveness of the proposed method on such datasets with an increasing number of examples.
Specifically, we consider creating an imbalanced dataset by increasing the number of examples in the head classes while keeping the number of examples in the tail classes constant. 
For example, in the case of Yahoo \citep{zhang2015character}, when the imbalance ratio $\phi = 1$, the number of annotated examples for each class remains 50.
When the imbalance ratio $\phi = 100$, the number of examples in the head classes is 5,000, but the number in the tail classes remains 50.
We report the test accuracy of four classification datasets with six existing selection methods.

Figure \ref{figure:4} (c) shows that imbalanced annotations significantly hurt ICL's performance when more examples are added to the head classes while keeping the number of examples in the tail classes constant.
This illustrates that the decreasing trend in ICL's performance mainly depends on the class prior distribution of the annotated dataset.
Notably, our method significantly improved the ICL's performance.
For instance, for a 100 imbalance ratio, our method outperforms vanilla by a large margin of 5.43, indicating the effectiveness of our method against imbalanced annotation.

\section{Conclusion}
In this paper, we presented a new research direction for understanding and mitigating the effect of imbalanced annotations on in-context learning (ICL). 
Our study reveals that imbalanced annotations significantly degrade the performance of ICL across various tasks, while classical rebalancing methods yield limited improvements.
We propose RCB, a complementary strategy that estimates conditional bias for reweighting the demonstration score in selection, universally enhancing the performance of ICL with imbalanced annotations.
To the best of our knowledge, our method is the first to tackle the class-imbalanced issue from the perspective of demonstration selection.
Extensive experiments demonstrate that RCB can enhance the performance of existing selection methods in ICL under class-imbalanced settings.
Our approach is easy to use in
practice, as it is insensitive to the hyperparameters and does not introduce heavy computational cost.

\textbf{Limitations.} Our methods need to select a balanced subset from imbalanced datasets, which might be impractical if there are few examples in the tail classes. 
It might be an interesting direction to explore how to model the bias in a more data-efficient manner.
\label{section:limitation}

% \section*{References}
\bibliographystyle{unsrt}
\bibliography{neurips_2025}

%%%%%%%%%%%%%%%%%%%%%%%%%%%%%%%%%%%%%%%%%%%%%%%%%%%%%%%%%%%%

\newpage
\appendix

\addcontentsline{toc}{section}{Appendix}
\renewcommand{\thepart}{} 
\renewcommand{\partname}{} 
\part{Appendix} % Start the appendix part
\parttoc % Insert the appendix TOC
% \tableofcontents

\section{Related Work}
\textbf{In-context learning} In-context learning (ICL) is a new paradigm for large language (LLMs), which allows LLMs to make predictions only based on a few demonstrations without explicitly updating parameters \citep{akyrek2023what,hendel2023context,agarwal2024many, dong2024survey,edwards-camacho-collados-2024-language,falck2024martingale}. 
Many studies show that ICL can achieve performance similar to fine-tuning but without the high computational cost \citep{gonen-etal-2023-demystifying, mosbach-etal-2023-shot,muller2024bayes,panwar2024incontext}. 
Despite achieving such outstanding performance, ICL has been criticized for being very sensitive to the quality of in-context examples \citep{fei-etal-2023-mitigating,gao2024noise}. 
Various approaches have been proposed to improve the robustness of ICL in recent years, including meta-tuning LLMs \citep{brunet2023icl}, calibration \citep{abbas2024enhancing},  demonstration selection \citep{zhang-etal-2022-active,nguyen2023context,qin2023context,ye-etal-2023-complementary,gao2024unifying,luo2024context,mo-etal-2024-c}, ordering \citep{lu-etal-2022-fantastically,liu2024let}, number \citep{zhang2025more} and formation \citep{voronov2024mind, yao-etal-2024-samples}.

The ICL is closely related to the concept of label bias \citep{brown2020language}, where language models are biased toward certain answers during few-shot learning.
In particular, the most relevant part is the majority label bias, a type of context label bias that leads the model to predict answers that appear frequently in the prompt \citep{zhao2021calibrate,gupta2023robust}. 
An empirical work \citep{wang-etal-2024-bayesian} suggested that label imbalance does not substantially impact ICL performance in binary classification. 
The subsequent works \citep{fei-etal-2023-mitigating,chu2023fine, hong2024mixtures, reif2024beyond} delved deeper into this challenge and proposed to address it by calibrating the model's output probabilities to compensate for this bias. 
While prior studies have typically focused on imbalanced distributions within prompts, our work shifts the emphasis to class imbalance in the annotation datasets from which demonstrations are drawn.
Notably, this study represents the first systematic attempt to tackle the class-imbalanced issue from the perspective of demonstration selection.

\textbf{Learning with imbalanced datasets} 
Imbalanced datasets are common in many real-world datasets, so the challenge of class imbalance has been widely studied in the literature \citep{cui2018large,cui2019class,jamal2020rethinking, schultheis2024generalized}. 
There are two popular directions in learning with imbalanced datasets: (1) Re-sampling: Re-sampling is a widely used strategy in class-imbalanced learning \citep{shi2023re}, such as Over-sampling \citep{chawla2002smote}, which involves repeating data from the minority classes; Under-sampling \citep{liu2008exploratory}, which involves removing a proportion of data from the majority classes.
Stratified sampling \citep{vilarino2005experiments} samples from each class have an identical probability of being sampled. 
In this work, we demonstrate that existing rebalancing methods yield limited improvement in in-context learning (ICL) when dealing with imbalanced annotations. 
(2) Training imbalance-robust models: designing loss function \citep{jamal2020rethinking,tan2020equalization,park2023robust,bhat2023robust,garcin2022stochastic} or designing model architectures \citep{long2022retrieval,pan2024lt} to mitigate the issue of imbalanced datasets. 
However, this method inevitably incurs high training costs that might be impractical for LLMs. In contrast, our approach only estimates the conditional bias for each class, resulting in negligible computational costs compared to traditional training methods.

\section{Proof}
\label{section:proof}
\textbf{\textit{Proof.}} We apply the importance sampling trick to connect the expected $\operatorname{Error}$ with the imbalanced annotated dataset:
\begin{align}
    \operatorname{Error}&=\mathbb{E}_{P_t(\mathrm{\mathbf{x}},\mathrm{\mathbf{y}})}M\left[ f_\theta\left( \operatorname{Top}_K \left( \left\{ s(\mathrm{\mathbf{c}}_i,\mathrm{\mathbf{x}}_t) \right\}_{i=1}^N \right), \mathrm{\mathbf{x}}_t\right), \mathrm{\mathbf{y}}_t \right], \nonumber \\
    &=\mathbb{E}_{P_c(\mathrm{\mathbf{x}},\mathrm{\mathbf{y}})}M\left[ f_\theta\left( \operatorname{Top}_K \left( \left\{ s(\mathrm{\mathbf{c}}_i,\mathrm{\mathbf{x}}_t) \right\}_{i=1}^N \right), \mathrm{\mathbf{x}}_t\right), \mathrm{\mathbf{y}}_t \right] \frac{P_t(\mathrm{\mathbf{x}},\mathrm{\mathbf{y}})}{P_c(\mathrm{\mathbf{x}},\mathrm{\mathbf{y}})}, \nonumber \\
    &=\mathbb{E}_{P_c(\mathrm{\mathbf{x}},\mathrm{\mathbf{y}})}M\left[ f_\theta\left( \operatorname{Top}_K \left( \left\{ s(\mathrm{\mathbf{c}}_i,\mathrm{\mathbf{x}}_t) \right\}_{i=1}^N \right), \mathrm{\mathbf{x}}_t\right), \mathrm{\mathbf{y}}_t \right] \frac{P_t(\mathrm{\mathbf{x}}|\mathrm{\mathbf{y}})P_t(\mathrm{\mathbf{y}})}{P_c(\mathrm{\mathbf{x}}|\mathrm{\mathbf{y}})P_c(\mathrm{\mathbf{y}})}, \nonumber 
\end{align}
Here, we decompose the term \(\frac{P_t(\mathrm{\mathbf{x}}|\mathrm{\mathbf{y}})P_t(\mathrm{\mathbf{y}})}{P_c(\mathrm{\mathbf{x}}|\mathrm{\mathbf{y}})P_c(\mathrm{\mathbf{y}})}\) into two parts: 
the class weights \(\bm{w} = P_t(\mathrm{\mathbf{y}})/P_c(\mathrm{\mathbf{y}})\), and the conditional weight \(\bm{\widetilde{\beta}} = \frac{P_t(\mathrm{\mathbf{x}}|\mathrm{\mathbf{y}})}{P_c(\mathrm{\mathbf{x}}|\mathrm{\mathbf{y}})}\) . The class weights measure the difference between class prior distributions $P(Y)$, and the conditional weight measures the difference between conditional distributions $P(X|Y)$. Then we have 
\begin{align}
    \operatorname{Error} =\mathbb{E}_{P_c(\mathrm{\mathbf{x}},\mathrm{\mathbf{y}})}M\left[ f_\theta\left( \operatorname{Top}_K \left( \left\{ s(\mathrm{\mathbf{c}}_i,\mathrm{\mathbf{x}}_t) \right\}_{i=1}^N \right), \mathrm{\mathbf{x}}_t\right), \mathrm{\mathbf{y}}_t \right] \bm{w} (1+\bm{\widetilde{\beta}}) \nonumber
\end{align}

where $\bm{w}$ denotes class weights and conditional weight $\bm{\widetilde{\beta}}= \frac{P_t(\mathrm{\mathbf{x}}|\mathrm{\mathbf{y}})}{P_c(\mathrm{\mathbf{x}}|\mathrm{\mathbf{y}})}-1$ measures the difference of conditional distribution between the annotated and test datasets.

For simplicity, we introduce a conditional bias $\bm{\beta} = \bm{w} \times \bm{\widetilde{\beta}}$  and re-write the expected $\operatorname{Error}$ as:
\begin{align}
    \mathbb{E}_{P_c(\mathrm{\mathbf{x}},\mathrm{\mathbf{y}})}M\left[ f_\theta\left( \operatorname{Top}_K \left( \left\{ s(\mathrm{\mathbf{c}}_i,\mathrm{\mathbf{x}}_t) \right\}_{i=1}^N \right), \mathrm{\mathbf{x}}_t\right), \mathrm{\mathbf{y}}_t \right](\bm{w}+\bm{\beta}). \nonumber
\end{align}
where conditional bias $\bm{\beta} = \bm{w} \times \bm{\widetilde{\beta}} = \frac{P_t(\mathrm{\mathbf{y}})}{P_c(\mathrm{\mathbf{y}})} \left( \frac{P_t(\mathrm{\mathbf{x}}|\mathrm{\mathbf{y}})}{P_c(\mathrm{\mathbf{x}}|\mathrm{\mathbf{y}})}-1 \right)$ denotes the deviation of class weights caused by the presence of conditional weight $\bm{\widetilde{\beta}}$. 

\section{Estimating  Conditional Bias by Bayesian optimization}
\label{bayeopt}
In this section, we will introduce how to use Bayesian optimization \citep{gardner2014bayesian,bayesianopt} to estimate conditional bias $\bm{\beta}$ as follows:
\begin{align}
\label{formula:10}
\bm{\beta} = \mathop{\arg\min}_{\bm{\beta}}
\frac{1}{|\mathcal{D}_b|} \sum_{i=1}^{|\mathcal{D}_b|} M\left[ f_\theta\left( \operatorname{Top}_K \left( \{(\bm{w}+\bm{\beta}) \times s(\mathrm{\mathbf{c}}_i,\mathrm{\mathbf{x}}_b) \right\}_{i=1}^{|\mathcal{D}_r|}), \mathrm{\mathbf{x}}_b\right), \mathrm{\mathbf{y}}_b \right],
\end{align}
\textbf{Surrogate Model.} We use the Gaussian Process as a surrogate model to approximate the objective function \ref{formula:10}. Initially, a prior distribution is established in a general form for the estimated model parameters,
\begin{align}
\mathcal{F}_\theta\left(\bm{\beta}\right) = \mathcal{N}\left( \mu(\bm{\beta}), \sigma^2(\bm{\beta}) \right), 
\end{align}
where $\mathcal{N}$ is the Gaussian distribution with a mean function $\mu(\bm{\beta})$ and a covariance function $\sigma^2(\bm{\beta})$. For a data point $\bm{\beta}_j$ sampled from the Gaussian process $\mathcal{N}$, we compute its corresponding function values $\mathcal{F}_\theta\left(\bm{\beta}_j\right)$.

\textbf{Acquisition Function.} We use the Expected Improvement (EI) criterion to select the point $\bm{\beta}_{j+1}$ with the maximum expected improvement as the next query point:
\begin{align}
    \operatorname{EI}(\bm{\beta}) & =\left\{\begin{array}{ll}
\left(\mu_{j}(\bm{\beta})-\mathcal{F}_\theta\left(\bm{\beta}^{+}\right)-\epsilon\right) \Phi(Z)+\sigma_{j}(\bm{\beta}) \phi(Z) & \text { if } \sigma_{j}(\bm{\beta})>0, \\
0 & \text { if } \sigma_{j}(\bm{\beta})=0.
\end{array}\right.  \\
Z & =\frac{\mu_{j}(\bm{\beta})-\mathcal{F}_\theta\left(\bm{\beta}^{+}\right)-\epsilon}{\sigma_{j}(\bm{\beta})}. 
\end{align}
where $\mu_{j}(\bm{\beta})$ and $\sigma_{j}(\bm{\beta})$ represent the mean and standard deviation of the surrogate model at step $j$, $\bm{\beta}^{+}$ denotes the current optimal observed point, $\epsilon$ is a tunable hyper-parameter. $\Phi(Z)$ and $\phi(Z)$ are the probability density function and cumulative density function of the Gaussian Process, respectively.

\textbf{Iterative Optimization Process.} We start with an initial set of observations $\bm{\beta}_0,...,\bm{\beta}_h$ by evaluating the objective function $\left\{\mathcal{F}\left(\bm{\beta}_i\right)\right\}_{i=0}^h$ at a few selected points. Second, we use the observed data $\left(\bm{\beta}_j, \mathcal{F}\left(\bm{\beta}_j\right)\right)$ to update the Gaussian Process model, refining the estimates of $\mu_{j}(\bm{\beta})$ and  $\sigma_{j}(\bm{\beta})$. Third, we employ the \textit{Expected Improvement} criterion to determine the next point $\bm{\beta}_{j+1}$ and obtain $\left\{\mathcal{F}\left(\bm{\beta}_{j+1}\right)\right\}_{i=0}^h$. We continue the process of updating the model and selecting new points until a stopping criterion is met. Finally, we select the point $\bm{\beta}^{+}$ with the best observed value of the objective function $\mathcal{F}_\theta\left(\bm{\beta}^{+}\right)$ as the optimal solution.

\section{Experimental Setting}
\label{appendix:experimental_setting}
\subsection{Datasets} 
\textbf{Datasets} We conduct experiments on various classification and generation tasks and examples of each dataset are shown in Tables \ref{table:a11} and \ref{table:a12}. 
We collect all datasets from Huggingface. 
The train sets of these datasets are regarded as example datasets and the test datasets are used to evaluate the performance of ICL. 
We randomly subsample examples from the test dataset to verify the performance of the imbalanced annotated dataset. 
We generate categories for Natural question \citep{kwiatkowski-etal-2019-natural} using ChatGPT-3.5-Turbo. The categories include people, time, geography, culture, and specialized knowledge.
\subsection{Baselines} 
\textbf{Baselines} To verify the effectiveness of our method, we compare our method with previous methods for demonstration retrieval by the downstream ICL's performance. 
We consider both learning-free and other learning-based retrievers as baselines, including
\begin{enumerate}
    \item \textbf{Random} selects demonstrations randomly from an example set without repetition \citep{min-etal-2022-metaicl}.
    \item \textbf{TopK} retrieves demonstrations that are semantically similar to a test query sample \citep{liu-etal-2022-makes}.
    \item \textbf{DPP} uses the original BERT embeddings as mentioned above without fine-tuning, and adopts MAP inference for subset retrieval \citep{Ye2023DPP}.
    \item \textbf{VoteK} proposes an unsupervised and graph-based selective annotation method to select diverse and representative demonstrations \citep{su2023selective}.
    \item  \textbf{ConE} searches for demonstrations by minimizing the difference in cross-entropy between the test input and the demonstrations \citep{peng-etal-2024-revisiting}.
    \item \textbf{ByCS} assumes that an accurate inverse likelihood probability will lead to an accurate posterior probability and selects demonstrations based on their inverse inference results \citep{wang-etal-2024-bayesian}.
\end{enumerate}

\subsection{Inference} 
\textbf{Perplexity} For classification tasks, we compute the sentence perplexity for each sequence formed by concatenating the input with each candidate answer \citep{brown2020language, wu-etal-2023-openicl}. Specifically, for each input instance $\mathrm{\mathbf{x}}$ is paired potential label set $\mathcal{Y}$, where $\mathcal{Y}$ represents the set of possible classes (e.g., Sports, Business, etc.). Then, for each possible label $\mathrm{\mathbf{y}}\in \mathcal{Y}$, we concatenate each tokenized input-output pair $(\mathrm{\mathbf{x}}, \mathrm{\mathbf{y}})$,  and obtain the corresponding tokenized sequence $\mathrm{\mathbf{c}}=(z_1,...,z_{|\mathrm{\mathbf{c}}|})=(x_1,...,x_{|\mathrm{\mathbf{x}}|},y_1,...,y_{|\mathrm{\mathbf{y}}|})$, where $|\mathrm{\mathbf{c}}|=|\mathrm{\mathbf{x}}|+|\mathrm{\mathbf{y}}|$. Now, the perplexity of $\mathrm{\mathbf{c}}$ is calculated as:
 \begin{equation}
\operatorname{Perplexity}(\mathrm{\mathbf{c}}) = \operatorname{exp}\{-\frac{1}{|\mathrm{\mathbf{c}}|}\sum^{|\mathrm{\mathbf{c}}|}_{i = 1} \log p_{\theta} (c_{i}|c_{< i})\}. \nonumber
\end{equation}
where $\log p_{\theta} (c_i|c_{< i})$ is the log-likelihood of the $i$-th token conditioned on the preceding tokens $c_{< i}$, from the given language model parameterized by $\theta$.  We select the label corresponding to the input-output pair $\mathrm{\mathbf{c}}$ with the lowest perplexity as the predicted label for the input $\mathrm{\mathbf{x}}$.

\textbf{Direct} For text generation tasks, we represent candidate answers using tokens from the vocabulary and select the final prediction based on the one with the highest probability \citep{brown2020language, wu-etal-2023-openicl}. To reduce computational cost, we set the maximum new tokens to 50 and limit unnecessary token generation.

\subsection{Experiment details} 
\textbf{Experiment details} We run our experiments on NVIDIA GeForce RTX 4090 and NVIDIA L40 GPU, and implement all methods by \textit{PyTorch} and \textit{transformers}.  Our code is inspired by OpenICL \cite{wu-etal-2023-openicl}. We thank the authors for releasing their code.
\label{section:gpu}

\section{More empirical results}

\subsection{Performance on different model architectures and sizes}
\label{Appendix_different_model}
We conduct experiments on various sized LLMs (including open-weight models and APIs). The results (vanilla/ours) in the table \ref{table:a5} below present the same phenomenon as the main experiments in the manuscript: the ICL performance of LLMs get worse at larger imbalanced ratios, and our method can significantly improve the performance, especially at large imbalanced ratios (e.g., 100).
\begin{table}[h]
\caption{Average accuracy (\%) of Vanilla / \textbf{+Ours} methods on the AgNews and Yahoo datasets with different model architectures and sizes. Bold numbers are superior results.}
    \centering
    \begin{tabular}{ccccc}
    \toprule
\textbf{Imbalance Ratio}&   1&\textbf{10}& \textbf{50} &\textbf{100}\\
\midrule
\multicolumn{5}{c}{Vanilla/\textbf{+Ours}}\\\hline
OPT-6.7B&  68.71/68.63&62.90/\textbf{67.54}& 57.35/\textbf{64.92}&54.31/\textbf{62.07}\\
OPT-13B&  72.22/72.84&68.04/\textbf{70.12}&62.28/\textbf{68.16}&58.12/\textbf{64.33}\\
OPT-30B&  76.34/76.60&72.32/\textbf{74.10}&67.74/\textbf{71.90}&64.30/\textbf{70.08}\\
LLAMA-3-8B&  78.90/79.10&75.20/\textbf{77.16}&69.32/\textbf{75.82}&65.72/\textbf{73.20}\\
LLAMA-3-70B&  86.50/86.43&83.07/\textbf{84.80}&79.86/\textbf{83.13}&77.60/\textbf{81.22}\\
ChatGPT-3.5-Turbo&   82.45/82.40&79.47/\textbf{81.08}& 77.88/\textbf{80.20}&76.32/\textbf{79.74}\\
Gemini-2.0-Flash&  81.91/82.09&79.95/\textbf{81.86}& 77.19/\textbf{80.31}&75.01/\textbf{79.82}\\ \hline
    \end{tabular}
    \label{table:a5}
\end{table}

\subsection{Can our method improve real-world imbalanced dataset}
\begin{figure}[h]
\centering % 确保整个figure*环境中的内容居中  
{\includegraphics[width=0.49\textwidth]{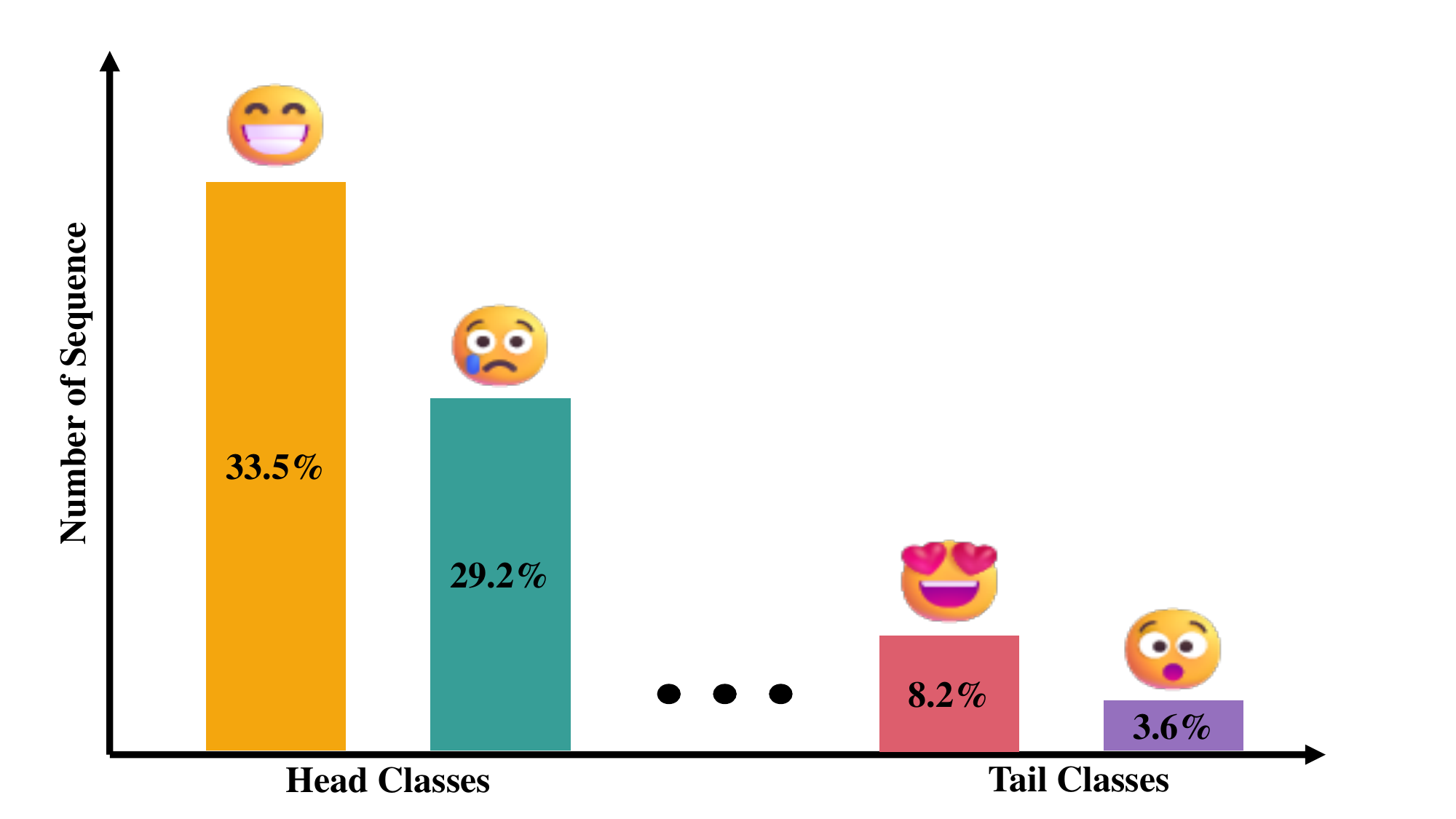}} % 
\caption{\textbf{An example of imbalanced dataset.} In Emotion \citep{saravia-etal-2018-carer},  a few sentiments make a large contribution and data tend to show a long-tailed distribution.  For example, ``joy`` and ``sadness`` are head classes, while most other classes, such as ``love`` and ``surprise``, are tail classes.}
\label{figure:a1}
\end{figure}

We verify the effectiveness of our method on a real-world imbalanced dataset. 
The Emotion \citep{saravia-etal-2018-carer} dataset has a long-tailed distribution (refer to Figure \ref{figure:a1}).  Table \ref{table:a8} shows that our method consistently improves existing selection methods and outperforms existing rebalancing methods, including over-sampling \citep{chawla2002smote}, under-sampling \citep{liu2008exploratory}, stratified sampling \citep{vilarino2005experiments}, and re-weighting \citep{cui2019class}. 
For example, with OPT-6.7B \citep{zhang2022opt}, using our method boosts the average Macro-F1 of six selection methods from 32.12 to 36.69, a direct improvement of 4.57. 
These results verify that our method is effective in improving ICL's performance in real-world imbalanced scenarios.
\begin{table}[h]
\caption{Average Macro-F1 metric on Emotion dataset. Bold numbers are superior results.}
    \centering
    \resizebox{1\columnwidth}{!}{%
    \begin{tabular}{ccccccc}
    \toprule
         \textbf{Methods}&  Vanilla& +Over-sampling&+Under-sampling&+Re-Weighting &+Stratified-Sampling &+Ours\\ \hline
    Accuracy& 32.22& 33.74&34.22&30.27&31.60&36.69\\ \hline
    \end{tabular}}
    \label{table:a8}
\end{table}

\subsection{Reweighting with conditional bias vs. output calibration?}
The ICL is closely related to the concept of label bias \citep{brown2020language}, where language models are biased toward certain answers during few-shot learning.
The subsequent works \citep{fei-etal-2023-mitigating,chu2023fine, hong2024mixtures, reif2024beyond} delved deeper into this challenge and proposed to address it by calibrating the model's output probabilities to compensate for this bias. 
One may also ask: \textit{can a similar effect be achieved by calibrating the output of ICL with imbalanced demonstrations?} 
In this ablation, we compare our method with some calibration methods, including CC \citep{zhao2021calibrate}, DC \citep{fei-etal-2023-mitigating} and LOOC \citep{reif2024beyond}.

Table \ref{table:a13} presents that although both the demonstration selection and calibration methods improve the performance of ICL against imbalanced annotations, the calibration methods obtain inferior performance compared to our proposed method. 
This is because the calibration methods \citep{zhao2021calibrate,fei-etal-2023-mitigating,reif2024beyond} only ensure that ICL with imbalanced demonstrations generates consistent output with that of balanced demonstrations. However, some effective selection methods, such as TopK \citep{liu-etal-2022-makes} and DPP \citep{ye-etal-2023-complementary}, are more likely to select examples belonging to the same class as the test sample for demonstrations and conduct imbalanced demonstrations. For these selection methods, the calibration methods inevitably degrade the performance of ICL, whereas our method can preserve their effectiveness.
\begin{table}[h]
\caption{Average test accuracy ($\%$) on AgNews dataset with various imbalanced ratios. Bold numbers are superior results.}
    \centering
    \begin{tabular}{ccccc}
    \toprule
         Imbalance Ratios&   1&10& 50&100\\ \hline
    Vanilla&  84.45&75.19& 67.78&64.02\\
 +CC& 78.38 & 77.29 & 75.75 &74.42 
\\
 +DD& 79.63 & 78.67 & 76.71 &75.58 
\\
 +LOOC& 75.71 & 73.00 & 74.25 &73.88 
\\
\textbf{ +Ours}&  \textbf{84.26}&\textbf{83.35}& \textbf{80.28}& \textbf{75.28}\\ \hline
    \end{tabular}
    \label{table:a13}
\end{table}

\subsection{Discussion of imbalanced test set}
We also conducted experiments on an imbalanced test set. The table below presents the average Macro-F1 metric on four datasets: AgNews, Amazon, Yelp, and Yahoo. The results in Table \ref{table:a9} demonstrate that our method can improve ICL performance on an imbalanced test set. 
For example, on the AgNews and Yahoo datasets with 100 imbalance ratio, our approach improves the test Macro-F1 of vanilla ICL from 45.42 to 53.85 – a 8.43 of direct improvement.

\begin{table}[h]
\caption{Average Macro-F1 metric of Vanilla / \textbf{+Ours} methods on the AgNews and Yahoo datasets with imbalanced test set. Bold numbers are superior results.}
    \centering
    \begin{tabular}{ccccc}
    \toprule
         Imbalance Ratios&   1&10& 50&100\\ \hline
    Vanilla&  64.32&58.11& 54.09&45.42
\\
 +Ours&  \textbf{64.52}&\textbf{61.67}& \textbf{57.89}& \textbf{53.85}\\ \hline
    \end{tabular}
    \label{table:a9}
\end{table}

\subsection{Results of extreme imbalanced ratios}
We conduct experiments on datasets with larger imbalanced ratios (e.g., 500, 1000). 
The table \ref{table:a6} presents the average accuracy ($\%$) on AgNews and Yahoo datasets. The results show that our method works well in the cases of extreme imbalanced ratios (e.g. 1000).
Notably, the improvements are more significant in cases with larger imbalanced ratios.
For instance, the improvement of our method increases from 7.76\% to 11.81\% when the imbalanced ratio increases from 100 to 1000.
\begin{table}[h]
\caption{Average accuracy (\%) of Vanilla / \textbf{+Ours} methods on the AgNews and Yahoo datasets with extreme imbalance ratios. Bold numbers are superior results.}
    \centering
    \begin{tabular}{ccccc}
    \toprule
         Imbalance Ratios&   10&100& 500&1000
\\ \hline
    Vanilla&  62.90&54.31& 46.80&44.52
\\ 
 +Ours&  \textbf{67.54}&\textbf{62.07}& \textbf{58.34}& \textbf{56.33}
\\
 Improvement&  4.64&7.76
& 11.54& 11.81
\\ \hline
    \end{tabular}
    \label{table:a6}
\end{table}

\subsection{The computational cost of our method}
\label{section:computational_cost}
In this section, we illustrate the high efficiency of our method from the following aspects.
Firstly, our method estimates the conditional bias by accessing low-cost model outputs and does not require extensive hyperparameter tuning.
The table \ref{table:a7} reports the average computational cost (in minutes) between our method and parameter-efficient fine-tuning (PEFT) \citep{mangrulkar2022peft} on the AgNews dataset across various imbalanced ratios. For example, with an imbalanced ratio of 100, one iteration of Bayesian optimization can be completed in 2.51 minutes using OPT-6.7B on a single NVIDIA L40 GPU.

Besides, the bayesian optimization employed in our method is well-suited for efficiently optimizing non-differentiable and black-box functions with a few iterations \citep{gardner2014bayesian,sabbatella2023bayesian,agarwalsearching}. In fact, our method can achieve significant improvement with minimal computational cost using 30 iterations and a balanced subset of size 100. 
Our method is acceptable in practice and only model weights $w+\beta$ one time for the whole test dataset. 

The detailed settings of PEFT are as follows: Learning Rate (1e-5); LR Scheduler Type (cosine); Batch Size (8); Number of Training Epochs (5); Number of Warmup Steps (100); Weight Decay (0.01); Lora Alpha (32); Lora Dropout (0.05).

\begin{table}[ht]
\caption{Comparison of average computational cost (minute)  per iteration using RCB and PEFT  on AgNews.  Bold numbers are superior results.}
    \centering
    \begin{tabular}{ccccc}
    \toprule
         Imbalance Ratios&  1& 10&50&100\\ \hline
 PEFT& 72.35& 35.25& 27.52&24.38\\ 
    RCB& \textbf{3.04}& \textbf{2.58}&\textbf{2.53}&\textbf{2.51}\\ \hline
    \end{tabular}
    \label{table:a7}
\end{table}

\subsection{Main results for each dataset}
\label{appendix_more_results}
Tables \ref{table:a21}, \ref{table:a22} \ref{table:a23} and \ref{table:a24} report average test accuracy ($\%$) with standard deviation on AgNews, Yahoo, Amazon and Yelp datasets with various imbalanced ratios (over 3 runs), respectively.

\begin{table*}[h]
\caption{
Average test accuracy ($\%$) with standard deviation on AgNews dataset with various imbalanced ratios (over 3 runs). The bold indicates the improved results by integrating our method.}
    \centering
    \resizebox{1\columnwidth}{!}{%
    \begin{tabular}{ccccccccc}
    \toprule
         $K$&  \multicolumn{4}{c}{8}&  \multicolumn{4}{c}{16}\\
\toprule
         \textbf{Imbalance Ratio}&  \textbf{1}&  \textbf{10}&  \textbf{50}&  \textbf{100} &  \textbf{1}&  \textbf{10}&  \textbf{50}& \textbf{100} \\
\midrule
Random&  68.17$\pm$0.12&  58.58$\pm$1.25&  48.75$\pm$1.81&  46.25$\pm$0.35&  70.58$\pm$0.59&  61.33$\pm$0.82&  50.75$\pm$1.62& 48.25$\pm$1.24\\
 \textbf{+Ours}& 68.08$\pm$0.72& \textbf{67.75$\pm$1.43}& \textbf{69.75$\pm$0.82}& \textbf{69.92$\pm$1.78}&68.83$\pm$0.42& \textbf{69.83$\pm$0.31}& \textbf{70.01$\pm$2.36}&\textbf{73.42$\pm$4.19}
\\
 TopK& 86.67$\pm$0.12& 76.42$\pm$0.77& 70.25$\pm$1.27& 66.67$\pm$1.89& 88.75$\pm$0.00& 79.17$\pm$0.42& 71.50$\pm$1.08&68.00$\pm$0.35\\
\textbf{+Ours}&85.92$\pm$1.18& \textbf{86.17$\pm$0.62}& \textbf{81.83$\pm$1.53}& \textbf{76.50$\pm$1.24}& 88.33$\pm$0.24& \textbf{86.58$\pm$1.23}& \textbf{83.25$\pm$1.27}&\textbf{76.42$\pm$1.01}
\\
 DPP& 86.50$\pm$0.00& 76.42$\pm$1.83& 69.67$\pm$0.82& 66.25$\pm$1.34& 87.25$\pm$0.00& 78.42$\pm$1.36& 71.58$\pm$1.71&68.92$\pm$0.31
\\
\textbf{+Ours}& 86.42$\pm$0.12& \textbf{87.75$\pm$1.43}& \textbf{81.25$\pm$1.67}& \textbf{78.17$\pm$1.33}& 87.42$\pm$0.12& \textbf{87.42$\pm$1.84}& \textbf{82.83$\pm$1.53}&\textbf{77.50$\pm$2.30}\\
 VoteK& 85.17$\pm$0.12& 76.75$\pm$1.54& 69.58$\pm$0.82& 66.42$\pm$1.66& 86.42$\pm$0.12& 78.83$\pm$0.42& 72.17$\pm$0.92&68.67$\pm$1.23
\\
  \textbf{+Ours}& \textbf{85.50$\pm$0.20}& \textbf{85.92$\pm$0.24}& \textbf{82.17$\pm$1.71}& \textbf{77.42$\pm$1.50}&86.58$\pm$0.31& \textbf{86.25$\pm$1.41}& \textbf{82.50$\pm$0.00}&\textbf{76.75$\pm$1.41}
\\
 ConE& 85.58$\pm$1.12& 76.00$\pm$2.16& 69.00$\pm$1.97& 65.50$\pm$1.27& 85.75$\pm$1.14& 77.33$\pm$1.25& 71.08$\pm$0.82&66.5$\pm$0.54
\\
 \textbf{+Ours}& \textbf{85.83$\pm$0.94}& \textbf{82.83$\pm$1.31}& \textbf{79.50$\pm$0.54}& \textbf{74.85$\pm$2.46}& 86.08$\pm$0.72& \textbf{83.75$\pm$1.08}& \textbf{81.58$\pm$0.62}&\textbf{74.83$\pm$1.36}
\\
 ByDC& 85.40$\pm$0.43& 74.00$\pm$1.08& 67.58$\pm$1.16& 63.58$\pm$1.36& 87.92$\pm$0.47& 76.08$\pm$0.59& 69.58$\pm$1.12&63.75$\pm$0.54
\\
\textbf{+Ours}&  84.73$\pm$0.38&  \textbf{82.75$\pm$1.24}& \textbf{80.92$\pm$1.36}&  \textbf{77.17$\pm$1.85}&  \textbf{88.33$\pm$0.12}&  \textbf{86.25$\pm$1.41}& \textbf{81.50$\pm$1.41}& \textbf{72.75$\pm$1.41}
\\\hline Vanilla&82.92$\pm$0.32&73.03$\pm$1.44&65.81$\pm$1.31&62.45$\pm$1.31&84.45$\pm$0.39&75.19$\pm$0.81&67.78$\pm$1.21& 64.02$\pm$0.70\\
\textbf{+Ours}&82.75$\pm$0.59&\textbf{82.20$\pm$1.05}&\textbf{79.24$\pm$1.27}&\textbf{75.67$\pm$1.69}&84.26$\pm$0.32&\textbf{83.35$\pm$1.21}&\textbf{80.28$\pm$1.20}& \textbf{75.28$\pm$1.95}\\\hline
    \end{tabular}
    }
    \label{table:a21}
\end{table*}

\begin{table*}[h]
\caption{
Average test accuracy ($\%$) with standard deviation on Yahoo dataset with various imbalanced ratios (over 3 runs). The bold indicates the improved results by integrating our method.}
    \centering
    \resizebox{1\columnwidth}{!}{%
    \begin{tabular}{ccccccccc}
    \toprule
         $K$&  \multicolumn{4}{c}{8}&  \multicolumn{4}{c}{16}\\
\toprule
         \textbf{Imbalance Ratio}&  \textbf{1}&  \textbf{10}&  \textbf{50}&  \textbf{100} &  \textbf{1}&  \textbf{10}&  \textbf{50}& \textbf{100} \\
\midrule
Random&  
37.93$\pm$1.91&  39.53$\pm$2.22&  36.07$\pm$0.41&  33.13$\pm$1.47&  44.20$\pm$0.59&  44.47$\pm$0.81&  39.27$\pm$1.18& 37.60$\pm$0.59\\
 \textbf{+Ours}& 37.87$\pm$1.84& 38.87$\pm$0.81& 36.00$\pm$0.71& \textbf{37.73$\pm$1.64}&43.20$\pm$1.18& 42.93$\pm$0.25& \textbf{44.33$\pm$0.74}&\textbf{44.00$\pm$1.93}\\
 TopK& 52.80$\pm$0.00& 51.47$\pm$0.34& 45.53$\pm$0.68& 43.07$\pm$1.46& 53.67$\pm$0.19& 52.93$\pm$1.27& 50.27$\pm$0.90&47.27$\pm$1.32
\\
\textbf{+Ours}&\textbf{53.47$\pm$0.47}& \textbf{53.00$\pm$0.33}& \textbf{48.33$\pm$1.54}& \textbf{47.87$\pm$2.62}& \textbf{54.40$\pm$0.85}& \textbf{56.00$\pm$1.50}& \textbf{50.53$\pm$1.33}&\textbf{50.20$\pm$2.70}\\
 DPP& 
54.00$\pm$0.00& 52.80$\pm$0.16& 47.53$\pm$1.24& 44.47$\pm$2.00& 56.57$\pm$0.26& 54.03$\pm$0.97& 50.47$\pm$0.57&48.87$\pm$1.25
\\
\textbf{+Ours}& 53.80$\pm$0.28& 51.04$\pm$1.49& 48.13$\pm$0.74& \textbf{47.80$\pm$2.27}& 56.07$\pm$0.19& \textbf{54.87$\pm$0.68}& \textbf{51.07$\pm$1.46}&\textbf{51.60$\pm$1.82}\\
 VoteK& 52.13$\pm$0.19& 50.80$\pm$0.59& 46.67$\pm$1.09& 44.67$\pm$1.31& 56.37$\pm$0.45& 53.10$\pm$0.54& 48.70$\pm$1.08&46.20$\pm$0.82\\
  
\textbf{+Ours}& 51.73$\pm$0.52& 50.24$\pm$0.45& \textbf{48.53$\pm$1.84}& \textbf{47.73$\pm$1.59}&56.03$\pm$0.26& \textbf{55.10$\pm$1.10}& \textbf{52.03$\pm$1.31}&\textbf{51.07$\pm$0.94}
\\
 ConE& 
51.53$\pm$0.52& 49.20$\pm$0.86& 44.60$\pm$1.30& 41.47$\pm$1.89& 51.90$\pm$0.80& 49.37$\pm$0.90& 45.33$\pm$0.68&42.77$\pm$1.47
\\
 \textbf{+Ours}& 51.67$\pm$0.66& \textbf{50.07$\pm$0.41}& \textbf{46.93$\pm$0.75}& \textbf{45.53$\pm$1.16}& \textbf{52.90$\pm$0.67}& \textbf{49.37$\pm$0.90}& \textbf{49.00$\pm$1.50}&\textbf{47.43$\pm$0.83}
\\
 ByDC& 
53.67$\pm$0.74& 49.60$\pm$0.85& 44.80$\pm$1.70& 42.17$\pm$1.68& 55.03$\pm$0.58& 49.77$\pm$1.11& 47.40$\pm$0.86&44.87$\pm$1.25
\\
\textbf{+Ours}&  53.40$\pm$0.43&  \textbf{52.33$\pm$0.90}& \textbf{48.47$\pm$1.65}&  \textbf{46.50$\pm$0.57}&  \textbf{55.37$\pm$0.68}&  \textbf{52.07$\pm$0.94}& \textbf{50.40$\pm$0.86}& \textbf{48.87$\pm$0.94}
\\\hline Vanilla&
50.34$\pm$0.56&48.90$\pm$0.84&44.20$\pm$1.07&41.50$\pm$1.64&52.96$\pm$0.48&50.61$\pm$0.93&46.91$\pm$0.88& 44.60$\pm$1.12\\
\textbf{+Ours}&50.32$\pm$0.70&\textbf{49.26$\pm$0.73}&\textbf{45.96$\pm$1.36}&\textbf{45.53$\pm$1.64}&53.00$\pm$0.64&\textbf{51.72$\pm$0.90}&\textbf{49.56$\pm$1.20}& \textbf{48.86$\pm$1.53}\\\hline    \end{tabular}
    }
    \label{table:a22}
\end{table*}

\begin{table*}[h]
\caption{
Average test accuracy ($\%$) with standard deviation on Amazon dataset with various imbalanced ratios (over 3 runs). The bold indicates the improved results by integrating our method.}
    \centering
    \resizebox{1\columnwidth}{!}{%
    \begin{tabular}{ccccccccc}
    \toprule
         $K$&  \multicolumn{4}{c}{8}&  \multicolumn{4}{c}{16}\\
\toprule
         \textbf{Imbalance Ratio}&  \textbf{1}&  \textbf{10}&  \textbf{50}&  \textbf{100} &  \textbf{1}&  \textbf{10}&  \textbf{50}& \textbf{100} \\
\midrule
Random&  42.07$\pm$0.69&  36.97$\pm$0.64&  35.93$\pm$0.71&  35.47$\pm$0.38&  43.67$\pm$0.44&  38.70$\pm$0.60&  36.27$\pm$0.91& 35.57$\pm$0.09
\\
 \textbf{+Ours}& 41.63$\pm$0.89& \textbf{41.67$\pm$1.96}& \textbf{38.90$\pm$1.00}& \textbf{40.17$\pm$0.98}&43.00$\pm$0.60& \textbf{43.60$\pm$1.47}& \textbf{42.93$\pm$0.56}&\textbf{43.33$\pm$1.24}
\\
 TopK& 46.50$\pm$0.00& 42.43$\pm$1.02& 37.17$\pm$1.36& 37.13$\pm$1.31& 47.60$\pm$0.00& 43.20$\pm$0.27& 38.07$\pm$1.29&37.17$\pm$1.36
\\
\textbf{+Ours}&46.13$\pm$0.49& \textbf{45.27$\pm$0.38}& \textbf{42.47$\pm$1.29}& \textbf{41.23$\pm$0.36}& 47.30$\pm$0.40& \textbf{46.57$\pm$0.58}& \textbf{43.60$\pm$0.27}&\textbf{40.87$\pm$1.22}
\\
 DPP& 46.70$\pm$0.00& 41.53$\pm$0.36& 38.30$\pm$0.20& 37.73$\pm$0.64& 48.10$\pm$0.00& 43.37$\pm$0.18& 39.27$\pm$0.56&37.77$\pm$0.58
\\
\textbf{+Ours}& 46.30$\pm$0.27& \textbf{43.63$\pm$1.11}& \textbf{41.23$\pm$1.96}& \textbf{39.77$\pm$0.56}& 47.97$\pm$0.09& \textbf{45.47$\pm$0.44}& \textbf{42.13$\pm$0.96}&\textbf{40.23$\pm$0.44}
\\
 VoteK& 46.87$\pm$0.05& 40.90$\pm$0.64& 37.00$\pm$1.39& 37.03$\pm$1.56& 48.03$\pm$0.04& 43.50$\pm$0.53& 37.57$\pm$0.62&36.73$\pm$0.98
\\
  \textbf{+Ours}& 46.67$\pm$0.16& \textbf{43.23$\pm$1.49}& \textbf{42.50$\pm$1.00}& \textbf{40.57$\pm$0.64}&47.93$\pm$0.16& \textbf{44.83$\pm$0.58}& \textbf{43.17$\pm$0.69}&\textbf{40.73$\pm$0.64}
\\
 ConE& 46.13$\pm$0.38& 40.27$\pm$0.66& 36.60$\pm$0.86& 35.67$\pm$1.67& 48.10$\pm$0.07& 42.77$\pm$0.16& 37.87$\pm$0.76&36.33$\pm$0.91
\\
 \textbf{+Ours}& 46.33$\pm$0.44& \textbf{41.60$\pm$1.20}& \textbf{40.80$\pm$0.93}& \textbf{39.23$\pm$0.91}& 48.00$\pm$0.13& \textbf{44.57$\pm$0.84}& \textbf{42.93$\pm$0.62}&\textbf{39.87$\pm$0.42}
\\
 ByDC& 46.73$\pm$0.09& 41.60$\pm$1.40& 37.47$\pm$0.93& 36.57$\pm$1.10& 48.23$\pm$0.09& 43.43$\pm$0.11& 38.60$\pm$1.00&37.43$\pm$0.71
\\
\textbf{+Ours}&  47.07$\pm$0.49&  \textbf{44.27$\pm$0.58}& \textbf{41.37$\pm$1.58}&  \textbf{40.03$\pm$1.09}&  48.03$\pm$0.24&  \textbf{45.47$\pm$0.96}& \textbf{42.67$\pm$0.24}& \textbf{40.57$\pm$0.51}
\\\hline Vanilla&45.83$\pm$0.20&40.62$\pm$0.79&37.08$\pm$0.91&36.60$\pm$1.11&47.29$\pm$0.11&42.49$\pm$0.31&37.94$\pm$0.86& 36.83$\pm$0.77
\\
\textbf{+Ours}&
45.69$\pm$0.46&\textbf{43.28$\pm$1.12}&\textbf{41.21$\pm$1.29}&\textbf{40.17$\pm$0.76}&47.04$\pm$0.27&\textbf{45.08$\pm$0.81}&\textbf{42.91$\pm$0.56}& \textbf{40.93$\pm$0.75}
\\\hline
    \end{tabular}
    }
    \label{table:a23}
\end{table*}

\begin{table*}[h]
\caption{
Average test accuracy ($\%$) with standard deviation on Yelp dataset with various imbalanced ratios (over 3 runs). The bold indicates the improved results by integrating our method.}
    \centering
    \resizebox{1\columnwidth}{!}{%
    \begin{tabular}{ccccccccc}
    \toprule
         $K$&  \multicolumn{4}{c}{8}&  \multicolumn{4}{c}{16}\\
\toprule
         \textbf{Imbalance Ratio}&  \textbf{1}&  \textbf{10}&  \textbf{50}&  \textbf{100} &  \textbf{1}&  \textbf{10}&  \textbf{50}& \textbf{100} \\
\midrule
Random&  
42.63$\pm$0.38&  39.80$\pm$0.93&  37.60$\pm$0.73&  37.20$\pm$0.13&  44.27$\pm$0.36&  42.10$\pm$0.67&  38.90$\pm$0.27& 38.17$\pm$0.22
\\
 \textbf{+Ours}& 42.90$\pm$0.80& \textbf{42.97$\pm$1.42}& \textbf{41.40$\pm$1.07}& \textbf{42.20$\pm$0.13}&\textbf{45.40$\pm$0.60}& \textbf{45.63$\pm$0.36}& \textbf{44.40$\pm$1.73}&\textbf{43.30$\pm$1.40}\\
 TopK& 47.30$\pm$0.00& 45.73$\pm$0.58& 41.13$\pm$0.64& 40.27$\pm$1.36& 49.70$\pm$0.00& 46.97$\pm$0.38& 43.30$\pm$0.27&42.17$\pm$0.49
\\
\textbf{+Ours}&46.97$\pm$0.44& 45.40$\pm$0.67& \textbf{43.93$\pm$0.89}& \textbf{42.07$\pm$0.64}& 48.10$\pm$0.00& 46.60$\pm$0.40& \textbf{44.93$\pm$0.11}&\textbf{43.83$\pm$0.36}
\\
 DPP& 
46.80$\pm$0.00& 44.83$\pm$0.82& 41.97$\pm$0.84& 40.57$\pm$0.44& 48.10$\pm$0.00& 46.33$\pm$0.89& 43.90$\pm$0.13&42.77$\pm$0.18
\\
\textbf{+Ours}& 46.27$\pm$0.71& 44.60$\pm$0.33& \textbf{42.37$\pm$0.91}& \textbf{41.97$\pm$0.36}& 48.13$\pm$0.04& 45.83$\pm$0.78& \textbf{44.20$\pm$0.47}&\textbf{44.03$\pm$0.49}
\\
 VoteK& 48.10$\pm$0.20& 45.60$\pm$0.33& 41.77$\pm$0.36& 40.73$\pm$0.49& 49.20$\pm$0.00& 46.70$\pm$0.47& 43.40$\pm$0.53&42.07$\pm$0.56
\\
  
\textbf{+Ours}& 47.77$\pm$0.31& 45.40$\pm$0.53& 42.60$\pm$0.80& 41.63$\pm$0.31&49.00$\pm$0.13& 46.73$\pm$0.16& 44.10$\pm$0.53&43.17$\pm$0.58
\\
 ConE& 
46.20$\pm$0.00& 44.17$\pm$0.69& 41.23$\pm$1.18& 39.70$\pm$0.47& 48.63$\pm$0.04& 46.40$\pm$0.53& 42.73$\pm$0.56&41.87$\pm$0.29
\\
 \textbf{+Ours}& 46.53$\pm$0.44& \textbf{44.50$\pm$0.73}& \textbf{42.57$\pm$0.58}& \textbf{40.87$\pm$0.64}& 48.43$\pm$0.11& \textbf{46.57$\pm$0.42}& \textbf{43.23$\pm$0.22}&\textbf{42.53$\pm$0.38}
\\
 ByDC& 
47.70$\pm$0.33& 45.27$\pm$0.16& 41.93$\pm$0.58& 40.80$\pm$0.53& 48.97$\pm$0.09& 46.37$\pm$0.36& 43.07$\pm$0.11&42.00$\pm$0.33\\
\textbf{+Ours}&  47.43$\pm$0.18&  \textbf{45.60$\pm$0.40}& \textbf{42.27$\pm$0.51}&  \textbf{42.10$\pm$0.20}&  48.37$\pm$0.09&  46.33$\pm$0.24& \textbf{43.57$\pm$0.29}& \textbf{43.57$\pm$0.29}
\\\hline Vanilla&
46.46$\pm$0.15&44.23$\pm$0.59&40.94$\pm$0.72&39.88$\pm$0.57&48.14$\pm$0.08&45.81$\pm$0.55&42.55$\pm$0.31& 41.51$\pm$0.34
\\
\textbf{+Ours}&46.31$\pm$0.48&\textbf{44.74$\pm$0.68}&\textbf{42.52$\pm$0.79}&\textbf{41.81$\pm$0.38}&47.91$\pm$0.16&\textbf{46.28$\pm$0.39}&\textbf{44.07$\pm$0.56}& \textbf{43.46$\pm$0.58}
\\\hline    \end{tabular}
    }
    \label{table:a24}
\end{table*}

\begin{table}[t]
\caption{The statistics, split and evaluation metrics of each dataset.}
    \centering
    \begin{tabular}{ccccc}
    \toprule
         Data&  Train Set&  test dataset&  Classes&Evaluation\\
    \midrule
         Amazon &  25000&  1000&  5&Accuracy\\
 AgNews & 20000&  1000& 4&Accuracy\\
         Yelp&  25000&  1000&  5&Accuracy\\
 Yahoo& 50000& 500& 10&Accuracy\\
 Emotion& 15758& 1974& 6&Macro-F1 \\
  Natural Question& 25000& 500& 5&Exact Match\\
 CodeSearchNet& 18000& 500& 6&Rouge\\
 \hline
\end{tabular}
    \label{table:a11}
\end{table}

\begin{table}[t]
\caption{The instructions, inference templates and example cases of tasks.}
    \centering
    \resizebox{1\columnwidth}{!}{%
    \begin{tabular}{ccc}
    \toprule
         Dataset&  Prompt&  Example\\
    \midrule
         Amazon&  \makecell[l]{Task Instruction: Sentiment of the sentence\\ Inference Verbalizer: Great, Good, Okay, Bad, Terrible?\\ Input: \textit{Question} \\Output: \textit{Answer}} &  \makecell[l]{Task Instruction: Sentiment of the sentence\\ Inference Verbalizer: Great, Good, Okay, Bad, Terrible? \\Input: why give me a date for a month then when its suppose to ship, \\its running late \\Output: Terrible}\\ \hline
 AgNews& \makecell[l]{Task Instruction: Text Classification Task\\ Inference Verbalizer: World, Sports, Business or Science New Topic?\\ Input: \textit{Question} \\Output: \textit{Answer}}&\makecell[l]{Task Instruction: Text Classification Task \\Inference Verbalizer: World, Sports, Business or Science New Topic?\\ Input: EBay Buys 25 Percent Stake in Craigslist Network By MAY WONG  \\  SAN JOSE, Calif. (AP) -- Online auctioneer eBay Inc... \\Output: Science}\\
         \hline
         Yelp& \makecell[l]{Task Instruction: Sentiment of the sentence\\ Inference Verbalizer: Great, Good, Okay, Bad, Terrible?\\ Input: \textit{Question} \\Output: \textit{Answer}} &  \makecell[l]{Task Instruction: Sentiment of the sentence\\ Inference Verbalizer: Great, Good, Okay, Bad, Terrible? \\Input: Awesome place!!! You must go and try all the services!!!! \\Output: Good}\\
         \hline
         Yahoo & \makecell[l]{Topic of the text:\\ Society $\&$ Culture, Science $\&$ Mathematics, \\Health, Education $\&$ Reference, Computers $\&$ Internet,\\ Sports, Business $\&$ Finance, Entertainment $\&$ Music, \\Family $\&$ Relationships, Politics $\&$ Government?\\ Input: \textit{Question} \\Output: \textit{Answer}} &  \makecell[l]{Topic of the text: Society $\&$ Culture, Science $\&$ Mathematics, \\Health, Education $\&$ Reference, Computers $\&$ Internet,\\ Sports, Business $\&$ Finance, Entertainment $\&$ Music, \\Family $\&$ Relationships, Politics $\&$ Government?\\ Input: what is god's kingdom that we are told to pray for? \\Output: Computers $\&$ Internet} \\
    \hline
 Emotion& \makecell[l]{Task Instruction: Sentiment of the sentence \\Inference Verbalizer: Sadness, Joy, Love, Anger, Fear, Surprise? \\Input: \textit{Question} Output: \textit{Answer}}&\makecell[l]{Task Instruction: Sentiment of the sentence Inference \\Verbalizer: Sadness, Joy, Love, Anger, Fear, Surprise?\\ Input: \textit{im feeling generous this week}\\ Output: \textit{Joy}}\\ \hline
 NQ &\makecell[l]{Question: \textit{Question} \\Answer: \textit{Answer}}&\makecell[l]{Question: \textit{who is the CEO of what's up} \\Answer: \textit{Jan Koum}}\\ \hline
 CodeSearchNet& \makecell[l]{Summarize the code.\\ Input: \textit{Input}\\ Output: \textit{Output}}&\makecell[l]{Summarize the code.\\ Input: "func NewMessage() Message $\{$\\
	return Message$\{$\\
		Context: context.Background(),\\
		Headers: map[string]string$\{$$\}$,\\
		Data:    render.Data$\{$$\}$,\\
		moot:    $\&$sync.RWMutex{},$\}$$\}$"  \\Output: \textit{NewMessage builds a new message.}}\\ \hline
    \end{tabular}}
    \label{table:a12}
\end{table}

\begin{figure}[h]
\centering % 确保整个figure*环境中的内容居中  
{\includegraphics[width=0.5\textwidth]{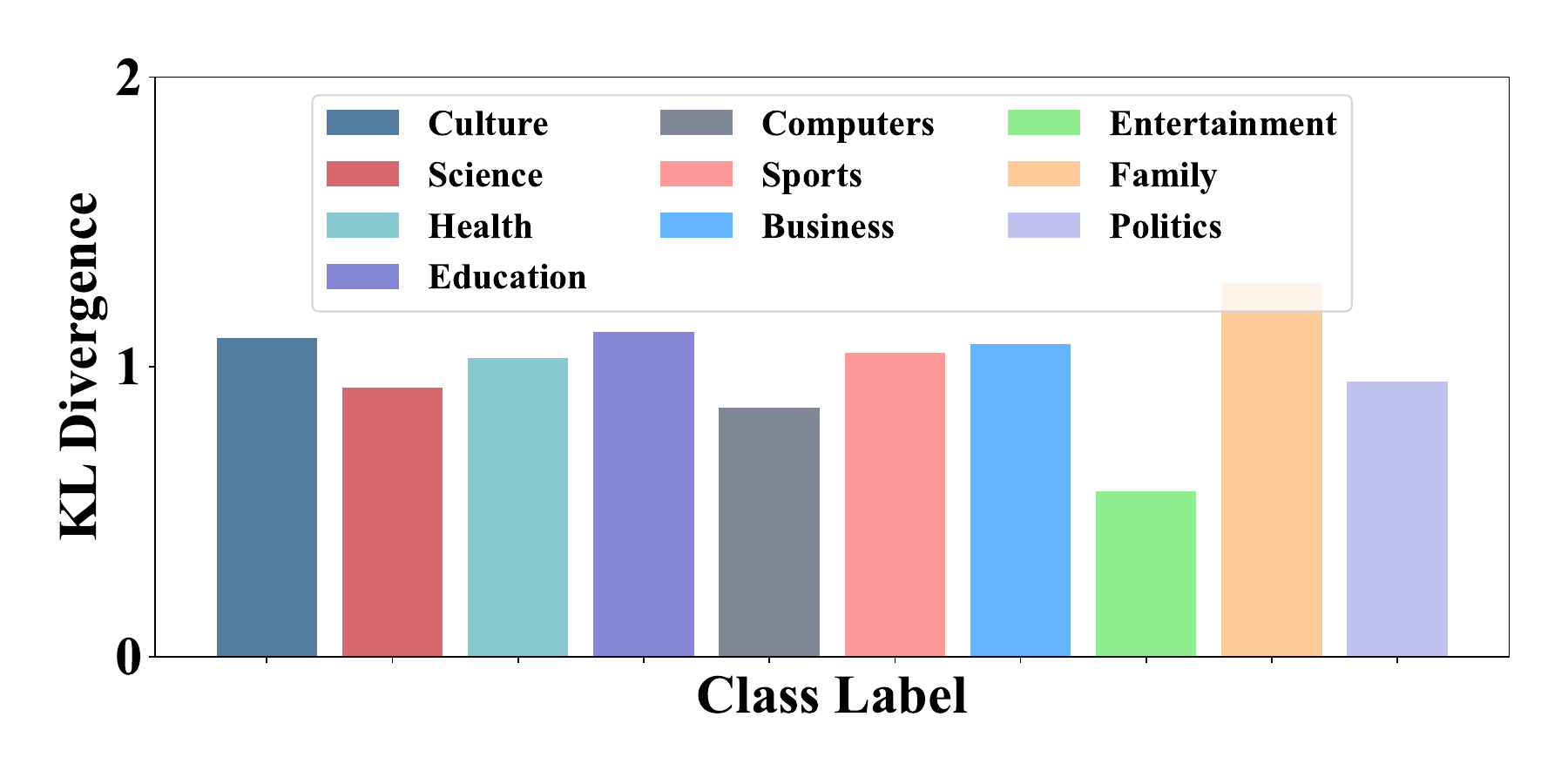}} % 
\caption{The Kullback-Leibler (KL) divergence between $P_c(X|Y)$ and  $P_t(X|Y)$ for each class in the Yahoo datasets. The KL divergence validates the existence of conditional bias in imbalanced datasets.}
\label{figure:a2}
\end{figure}

\begin{figure}[t]
\centering % 确保整个figure*环境中的内容居中  
\subfigure[Amazon]{\includegraphics[width=0.32\textwidth]{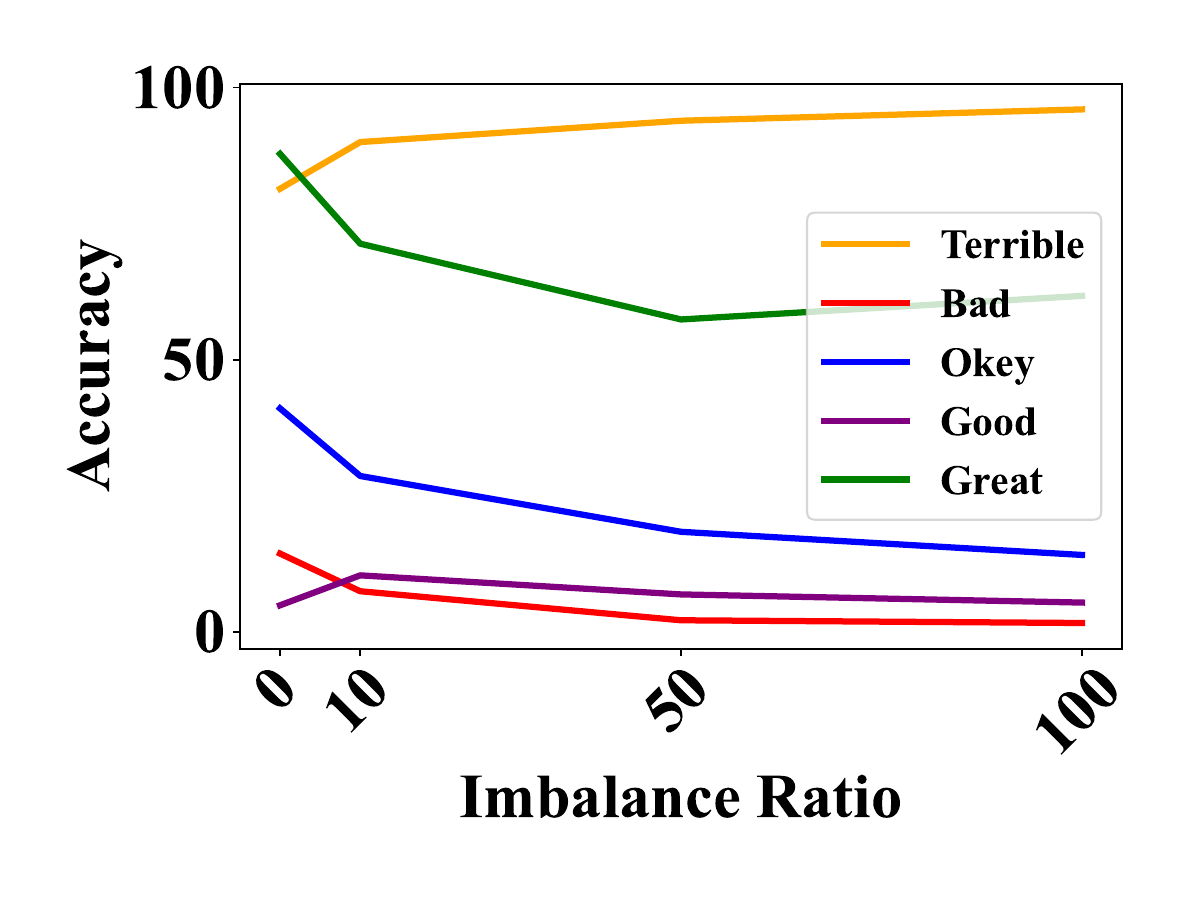}} %
\subfigure[Yelp]{\includegraphics[width=0.32\textwidth]{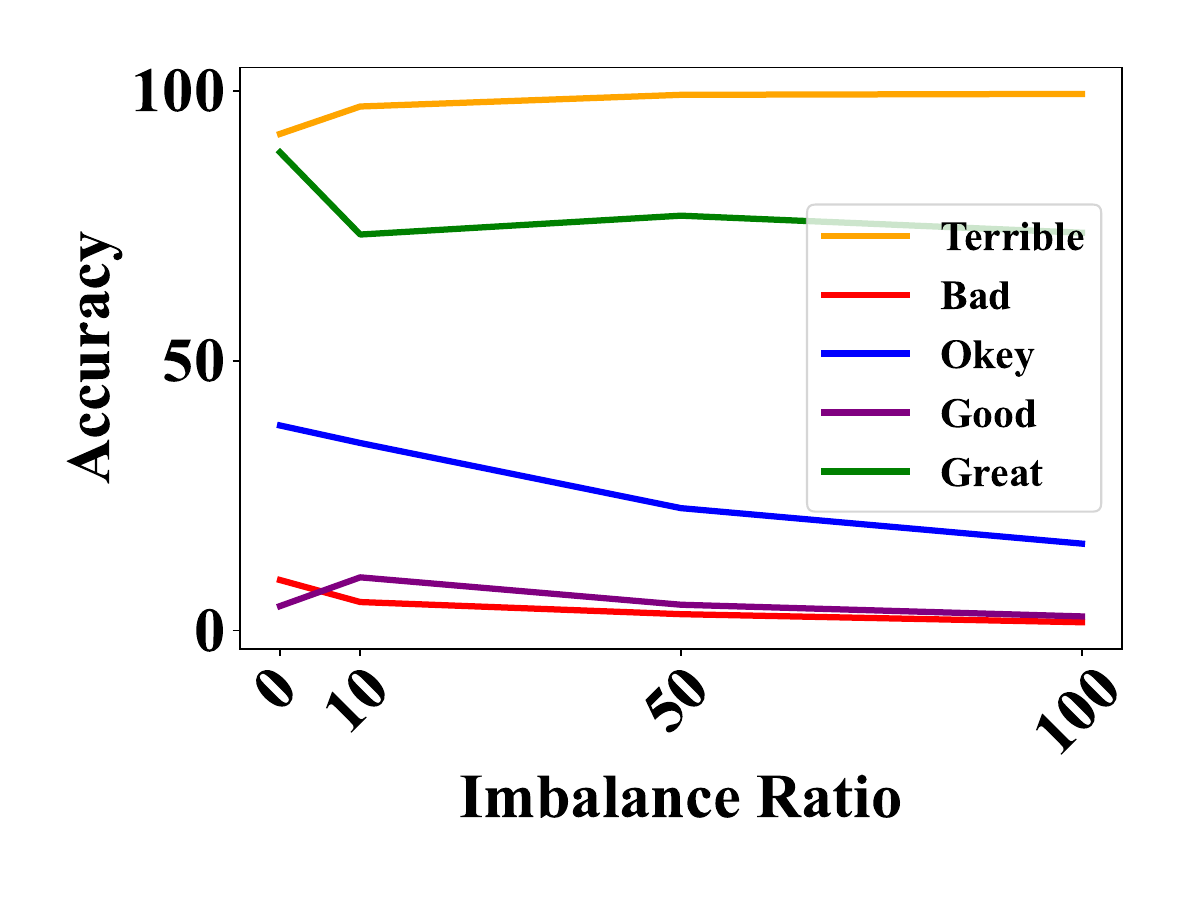}} % 
\subfigure[Yahoo]{\includegraphics[width=0.32\textwidth]{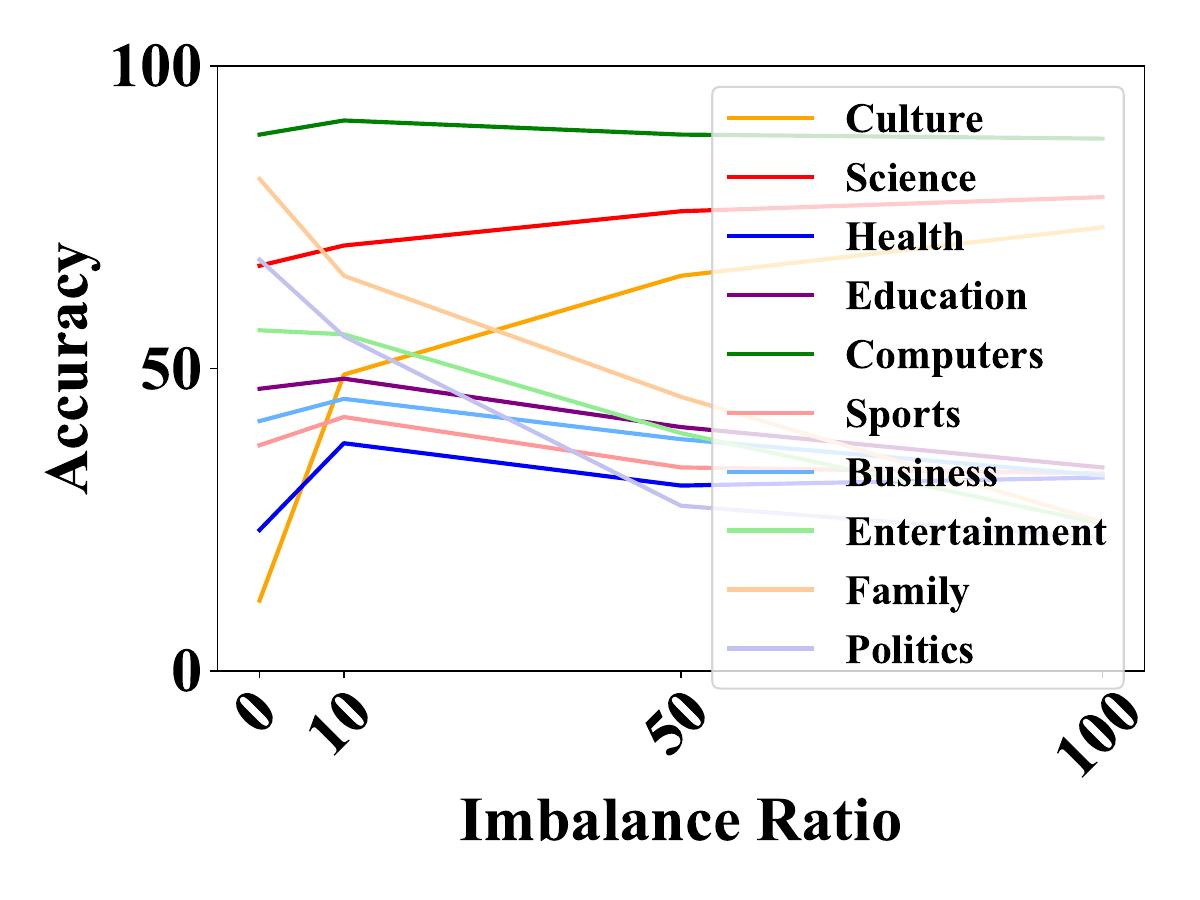}} %
\caption{The comparison of average accuracy for each class in the Amazon (a), Yelp (b) and Yahoo (c) datasets across various imbalance ratios.}
\label{figure:a3}
\end{figure}

\end{document}